\documentclass{article}

\usepackage[nonatbib, final]{neurips_2023}
\usepackage{cite}
\usepackage{stmaryrd}
\usepackage{textcomp}
\usepackage[mathscr]{euscript}
\usepackage{bbm}

\usepackage[inline]{enumitem}
\usepackage{graphicx}
\usepackage[font=footnotesize]{caption}
\usepackage{subcaption}
\usepackage[export]{adjustbox}
\usepackage{multirow}
\usepackage{multicol}
\usepackage[ruled, lined, linesnumbered, commentsnumbered, longend]{algorithm2e}
\makeatletter
\renewcommand{\fnum@algocf}{\AlCapSty{\AlCapFnt\thealgocf\nobreakspace\algorithmcfname}}%
\makeatother
\SetAlgorithmName{Pseudocode}{}{}

\makeatletter
\newcommand\notsotiny{\@setfontsize\notsotiny{6.5}{6}}
\makeatother

\usepackage[utf8]{inputenc} 
\usepackage[T1]{fontenc}    
\usepackage{hyperref}       
\usepackage{url}            
\usepackage{booktabs}       
\usepackage{amsfonts}       
\usepackage{nicefrac}       
\usepackage{microtype}      
\usepackage{xcolor}         
\usepackage{amssymb,amsmath,amsthm}

\title{On Comparing Fair Classifiers under Data Bias}

\author{%
  Mohit Sharma \thanks{Work partly done during an internship at Microsoft Research India. The author is supported by Microsoft Research India Joint PhD fellowship. \\ All correspondence to mohits@iiitd.ac.in.} \\
  \small{IIIT Delhi} \\
  \And
  Amit Deshpande \\
  \small{Microsoft Research India} \\
  \And
  Rajiv Ratn\\
  \small{IIIT Delhi}\\
}

\newtheorem{theorem}{Theorem}
\newtheorem{proposition}{Proposition}

\newtheorem{remark}{Remark}

\DeclareMathOperator*{\argmin}{\arg\!\min}
\begin{document}

\maketitle

\begin{abstract}
  In this paper, we consider a theoretical model for injecting data bias, namely, under-representation and label bias (Blum \& Stangl, 2019). We empirically study the effect of varying data biases on the accuracy and fairness of fair classifiers. Through extensive experiments on both synthetic and real-world datasets (e.g., Adult, German Credit, Bank Marketing, COMPAS), we empirically audit pre-, in-, and post-processing fair classifiers from standard fairness toolkits for their fairness and accuracy by injecting varying amounts of under-representation and label bias in their training data (but not the test data). Our main observations are: \begin{enumerate*} \item The fairness and accuracy of many standard fair classifiers degrade severely as the bias injected in their training data increases, \item A simple logistic regression model trained on the right data can often outperform, in both accuracy and fairness, most fair classifiers trained on biased training data, and \item A few, simple fairness techniques (e.g., reweighing, exponentiated gradients) seem to offer stable accuracy and fairness guarantees even when their training data is injected with under-representation and label bias\end{enumerate*}. Our experiments also show how to integrate a measure of data bias risk in the existing fairness dashboards for real-world deployments.
\end{abstract}

\section{Introduction}

Fairness in AI and machine learning is an important problem because automated decision-making in highly sensitive domains such as justice, healthcare, banking, and education can proliferate social and economic harms \cite{chouldechova2018frontiers, friedler2019comparative, barocas2017fairness, mehrabi2021survey}. Fair and representative data is a foremost requirement to train and validate fair models. However, most collected and labeled datasets invariably contain historical and systemic biases \cite{barocas2016big}, resulting in unfair and unreliable performance after deployment \cite{shankar2017no, buolamwini2018gender, wilson2019predictive}. Even many standard datasets used to evaluate and compare bias mitigation techniques are not perfectly fair and representative \cite{ding2021retiring,peng2021mitigating}.

Recent literature in fair classification has studied the failure of fairness and accuracy guarantees under distribution shifts (e.g., covariate shift, label shift) between the training and test distributions and proposed fixes under reasonable assumptions \cite{schumann2019transfer,dailabel, rezaei2020robust,singh2021fairness,maity2021does,du2021fair,biswas2021ensuring,schrouff2022maintaining}. In this paper, we consider a simple, natural data bias model that incorporates under-representation, over-representation, and label biases \cite{blum2019recovering}, and study the change in accuracy and fairness of various classifiers as we inject bias in their training data. The central question we ask is: \emph{Among the various fair classifiers available in the standard fairness toolkits (e.g., \cite{aif360-oct-2018}), which ones are less vulnerable to data bias?}

The definitions of fairness prevalent in literature fall into two broad types: individual fairness and group fairness \cite{dwork2012fairness,bechavod2020metric,petersen2021post,fleisher2021what,petersen2021post,hardt2016equality,zafar2017fairnessbeyond,binns2020}. Individual fairness promises similar treatment to similar individuals. Our focus is group-fair classification, which promises equal or near-equal outcomes across different sensitive groups (e.g., race, gender). Examples of group fairness include Equal Opportunity (equal false negative rates across groups) and Statistical Parity (equal acceptance rate across groups). Various bias mitigation techniques have been proposed for group-fair classification motivated by different applications and stages of the machine learning pipeline \cite{zliobaite2015survey, friedler2019comparative, mehrabi2021survey}. They are broadly categorized into three types: \begin{enumerate*}\item \emph{Pre-Processing}, where one transforms the input data, independent of the subsequent stages of training, and validation \cite{kamiran2012data, feldman2015certifying, calmon2017optimized}, \item \emph{In-Processing}, where models are trained by fairness-constrained optimization \cite{calders2010three,kamishima2012fairness,zemel2013learning,zafar2017fairness,zhang2018mitigating,agarwal2018reductions,celis2019classification}, and \item \emph{Post-Processing}, where the output predictions are adjusted afterward to improve fairness \cite{kamiran2012decision, hardt2016equality, pleiss2017fairness}\end{enumerate*}. These techniques appear side by side in standard fairness toolkits with little normative prescription about which technique to choose under data bias. Furthermore, it's unclear what happens to such standard fair classifiers when they are trained and exposed to varying amounts of data bias over time in the training or deployment environments. We attempt to close this gap by providing an auditing method to compare different fair classifiers under a simple model of data bias. 

To examine the effect of data bias on group-fair classifiers, we use a theoretical model by Blum \& Stangl \cite{blum2019recovering} for under-representation and label bias. It is a simple model for demographic under-/over-sampling and implicit label bias seen in real-world data \cite{shankar2017no, buolamwini2018gender, wilson2019predictive, greenwald2006implicit, kleinberg2018selection}. 
%
Given a train-test data split, we create biased training data by injecting under-representation and label bias in the training data (but not the test data). We then study the error rates and (un)fairness metrics of various pre-, in-, and post-processing fair classifiers on the original test data, as we vary the amount of bias in their training data. 
We inject two types of under-representation biases in the training data: $\beta_{\text{pos}}$-bias, where we undersample the positive labeled population of the minority group by a multiplicative factor of $\beta_{\text{pos}}$, and $\beta_{\text{neg}}$-bias, where we undersample the negative labeled population of the minority group by a multiplicative factor of $\beta_{\text{neg}}$. We examine the effect of increasing the parameters $\beta_{\text{pos}}$ and $\beta_{\text{neg}}$ separately as well as together. We also investigate the effect of label bias parameter $\nu$, where we flip the positive labels of the minority group to negative ones with probability $\nu$. We cannot hope to re-create real-world data bias scenarios via this synthetic pipeline \cite{buolamwini2018gender, wilson2019predictive}, but our findings on stability/instability of various classifiers can be somewhat useful towards designing fair pipelines in the presence of various data biases.

\paragraph{\bf Our Contributions:} Our main contributions and observations can be summarized as follows.
\begin{itemize}
    \item Using synthetic and real-world datasets, we show that the fairness and accuracy guarantees of many fair classifiers from standard fairness toolkits are highly vulnerable to under-representation and label bias. In fact, often, an unfair classifier (e.g., logistic regression classifier) trained on the correct data can be more accurate and fairer than fair classifiers trained on biased data. 
    \item Some fair classification techniques (viz., reweighing \cite{kamiran2012data}, exponentiated gradients \cite{agarwal2018reductions}, adaptive reweighing \cite{jiang2020identifying}) have stable accuracy and fairness guarantees even when their training data in injected with under-representation and label bias. We provide a theoretical justification to support the empirically observed stability of the reweighing classifier \cite{kamiran2012data}; see Theorem \ref{thm:sandwich}.
    \item Our experimental pipeline can be leveraged to create audit and test suites to check the vulnerability of fair classifiers to simple models of data bias before their real-world deployment.
\end{itemize}
The rest of the paper is organized as follows. Section \ref{sec:related-work} explains related work to set the context for our work. Section \ref{sec:exp-setup} explains our experimental setup, and Section \ref{sec:stability} discusses our key empirical observations. Section \ref{sec:theorems} contains some relevant theorems and proof outlines. Section \ref{sec:discussion} discusses our work in the context of other research directions on data bias in fair machine learning.

\section{Related Work} \label{sec:related-work}

The robustness of machine learning models when there is a mismatch between their training and test distributions has been studied under various theoretical models of distribution shifts, e.g., covariate shift and label shift. Recent works have studied fair classification subject to these distribution shifts and proposed solutions under reasonable assumptions on the data distribution \cite{coston2019fair, schumann2019transfer, blum2019recovering, rezaei2020robust, singh2021fairness, dailabel, biswas2021ensuring, du2021fair, maity2021does, giguere2022fairness}. As pointed out in the introduction, our goal is different, so we defer the critical contributions of these works to Appendix \ref{appndx:related_work_dist_shift}. However, later in Section \ref{sec:discussion}, we discuss different distribution shifts and their relation to the data bias model of Blum \& Stangl \cite{blum2019recovering} used in our paper.

Our work builds on the under-representation and label bias model proposed in Blum \& Stangl \cite{blum2019recovering}. Under this model, they prove that the optimal fair classifier that maximizes accuracy subject to fairness constraints (equal opportunity constraints, to be precise) on the biased data distribution gives the maximum accuracy on the unbiased data distribution. For under-representation bias, the above result can be achieved by equalized odds and equal opportunity constraints but not by demographic parity constraints. Along similar lines, Maity et al. \cite{maity2021does} derive necessary and sufficient conditions about when performance parity constraints in the presence of subpopulation shifts give better accuracy on the test distribution. Jiang et al. \cite{jiang2020identifying} have also considered the label bias model, although we look at the aggregate effect over varying amounts of label bias. Our observations on label bias also corroborate various findings by other works investigating the role of label noise in fair classification \cite{lamy2019noise, fogliato2020fairness, wang2021fair, konstantinov2022impossibility}. Under-representation bias can also be studied as a targeted Sample Selection Bias mechanism \cite{cortes2008sample}. There is some work done on fair sample selection bias \cite{du2021fair, zhu2023consistent}, which we elaborate on in Appendix \ref{appndx:related_work_dist_shift}.

Parallel to our work, a recent preprint by Akpinar et al. \cite{akpinar2022sandbox} proposes a simulation toolbox based on the under-representation setting from Blum \& Stangl \cite{blum2019recovering} to stress test four different fairness interventions with user-controlled synthetic data and data biases. On the other hand, our work uses the under-representation framework from Blum \& Stangl \cite{blum2019recovering} to extensively examine the stability of different types of fair classifiers on various synthetic and real-world datasets. While Akpinar et al. \cite{akpinar2022sandbox} focus on proposing a simulation framework to extensively test all findings of Blum \& Stangl \cite{blum2019recovering}, we focus on using the under-representation framework to highlight the differences in performance between various fair classifiers and theoretically investigating the stability of one of the fair classifiers.

We compare various fair classifiers in the presence of under-representation and label bias, which complements some expository surveys in fair machine learning \cite{zliobaite2015survey, mehrabi2021survey, pessach2022review, feng2022fair} and some benchmarking on important aspects like the choice of data processing, splits, data errors, and data pre-processing choices \cite{friedler2019comparative, biswas2021ensuring, qian2021my, islam2022through}. We elaborate on these works in the Appendix \ref{appndx:related_work_compare}.

\section{Experimental Setup} \label{sec:exp-setup}

For our analysis, we select a comprehensive suite of commonly used fair classifiers implemented in the open-source AI Fairness 360 (AIF-360) toolkit \cite{aif360-oct-2018}. From the available choice of pre-processing classifiers, we include two: \begin{enumerate*} \item Reweighing (`rew') \cite{kamiran2012data}, and \item Adaptive Reweighing (`jiang\_nachum') \cite{jiang2020identifying}\end{enumerate*}. Since `jiang\_nachum' is not a part of AIF-360, we use the authors' implementation\footnote{https://github.com/google-research/google-research/tree/master/label\_bias}. Among the in-processing classifiers in AIF-360, we choose three: \begin{enumerate*} \item Prejudice Remover classifier (`prej\_remover') \cite{kamishima2012fairness}, \item Exponentiated Gradient Reduction classifier (`exp\_grad'), and its deterministic version (`grid\_search') \cite{agarwal2018reductions} and \item the method from Kearns et al. (`gerry\_fair') \cite{kearns2018preventing} \end{enumerate*}. From post-processing classifiers, we choose three: \begin{enumerate*} \item the Reject Option classifier (`reject') \cite{kamiran2012decision}, \item the Equalized Odds classifier (`eq') \cite{hardt2016equality}, and \item the Calibrated Equalized Odds algorithm (`cal\_eq') \cite{pleiss2017fairness}\end{enumerate*}. We omitted the use of some classifiers in the AIF360 toolkit for two reasons: resource constraints and their extreme sensitivity to the choice of hyperparameters.

All fair classifiers use a base classifier, which it uses in its optimization routine to mitigate unfairness. We experiment with two base classifiers: a logistic regression (LR) model and a Linear SVM classifier with Platt's scaling \cite{platt1999probabilistic} for probability outputs. We use the Scikit-learn toolkit to implement the base classifiers \cite{pedregosa2011scikit}. The `prej\_remover' and `gerry\_fair' classifiers do not support SVM as a base classifier; hence, we only show results for those on LR.

The fairness metric we use in the main paper is Equalized Odds (EODDS) \cite{hardt2016equality}, which is defined as the difference of False positive and False negative Rates, and we measure an equally weighted combination of both to report our results $0.5*|\operatorname{Pr}(\hat{Y}=0 | Y=1, S=1) - \operatorname{Pr}(\hat{Y}=0 | Y=1, S=0)| + 0.5*|\operatorname{Pr}(\hat{Y}=1 | Y=0, S=1) - \operatorname{Pr}(\hat{Y}=1 | Y=0, S=0)|$, where $\hat{Y} \in \{0,1\}$ denotes the predicted label, $Y \in \{0,1\}$ denotes a binary class label, and $S \in \{0,1\}$ denotes a binary sensitive attribute. Blum \& Stangl \cite{blum2019recovering} suggest that equal opportunity and equalized odds constraints on biased data can be more beneficial than demographic parity constraints. Therefore, we use the equalized odds constraint for `jiang\_nachum' and `exp\_grad'. Different classifiers have been proposed for different fairness metrics and definitions. However, we often compare different methods with one or two standard metrics in practice. Since most of the methods used in this paper were trained for Equalized Odds, we report Equalized Odds results in the main text and report average Equal Opportunity Difference (EOD) and Statistical Parity Difference (SPD) results over under-representation bias settings to report a complete picture of the effect on data bias on fairness and error rate. EOD is defined as the True Positive Rate difference: $|\operatorname{Pr}(\hat{Y}=1 | Y=1, S=1) - \operatorname{Pr}(\hat{Y}=1 | Y=1, S=0)|$, while SPD is defined as $|\operatorname{Pr}(\hat{Y}=1 | S=1) - \operatorname{Pr}(\hat{Y}=1 | S=0)|$. For both metrics, values closer to $0$ mean better fairness, whereas values closer to $1$ indicate extreme parity difference between the two sensitive groups $S=0$ and $S=1$.


We train on multiple datasets to perform our analysis. We consider four standard real-world datasets from fair classification literature: the Adult Income dataset \cite{Dua:2019}, the Bank Marketing dataset \cite{moro2014data}, the COMPAS dataset \cite{angwin2016machine}, and the German Credit dataset \cite{Dua:2019}. We also consider a synthetic dataset setup from Zeng et al. \cite{zeng2022bayes}, which consists of a binary label $y \in \{0,1\}$ and a binary sensitive attribute setting $s \in \{0,1\}$. We take the original train-test split for a given dataset, or in its absence, a 70\%-30\% stratified split on the subgroups. Appendix \ref{appndx:dataset_info} gives specific details for all datasets. 

To perform our analysis, we use the data bias model from Blum \& Stangl \cite{blum2019recovering} and inject varying amounts of under-representation and label bias into the original training data before giving it to a classifier. We summarize our experimental setup in Algorithm $1$, presented in Appendix \ref{appndx:algorithm}. Let $Y \in \{0,1\}$ represent the label, and $S \in \{0,1\}$ represent the sensitive attribute. Let $\beta_{pos}$ be the probability of retaining samples from the subgroup defined by the favorable label $Y=1$ and underprivileged group $S=0$ ($10$-subgroup). Similarly, let $\beta_{neg}$ be the probability of retaining samples from the unfavorable label $Y=0$ and underprivileged $S=0$ group ($00$-subgroup). We inject under-representation bias into the training data by retaining samples from the $10$-subgroup and the $00$-subgroup with probability $\beta_{pos}$ and $\beta_{neg}$, respectively. We consider ten different values each for $\beta_{pos}$ and $\beta_{neg}$ varying over $\{0.1, 0.2, ..... , 1.0\}$. This results in $100$ different settings ($10~\beta_{pos}$ factors x $10~\beta_{neg}$ factors) for training data bias. We separately inject ten different levels of label bias by flipping the favorable label of the underprivileged group to an unfavorable one ($10 \rightarrow 00$) with a probability $\nu$ varying over $\{0.0, 0.1, ........, 0.9\}$. The test set for all settings is the original split test set, either provided at the original dataset source or taken out separately beforehand. This results in $110$ different datasets corresponding to different data bias settings, and consequently $110$ different results each for a particular fair classifier. Finally, the training of all classifiers and the procedures to create biased training data using  $\beta_{pos}$, $\beta_{neg}$, and $\nu$ is performed $5$ times to account for randomness in sampling and optimization procedures. 

We also note that some fair classifiers in our list, like `exp\_grad' and `cal\_eq', are randomized classifiers. Therefore, repeating the entire pipeline multiple times and reporting means and standard deviations normalizes the random behavior of these methods to some extent, apart from the measures taken during the implementation of these methods in widely used toolkits like the one we are using, AIF360 \cite{aif360-oct-2018}. The code to train and reproduce the results is available here \footnote{https://github.com/mohitsharma29/comparision\_data\_bias\_fairness}.


\section{Stability of Fair Classifiers: Under-Representation and Label Bias} \label{sec:stability}

In this section, we present our experimental results about the stability of fairness and accuracy guarantees of various fair classifiers after injecting under-representation and label bias in their training data. Stability means whether the error rate and unfairness show high variance/spread in response to increasing or decreasing under-representation or label bias. A stable behavior is indicated by small changes in error rate and unfairness, as $\beta_{pos}, \beta_{neg} \text{ or } \nu$ change.

\begin{figure}
     \centering
     \includegraphics[width=\textwidth]{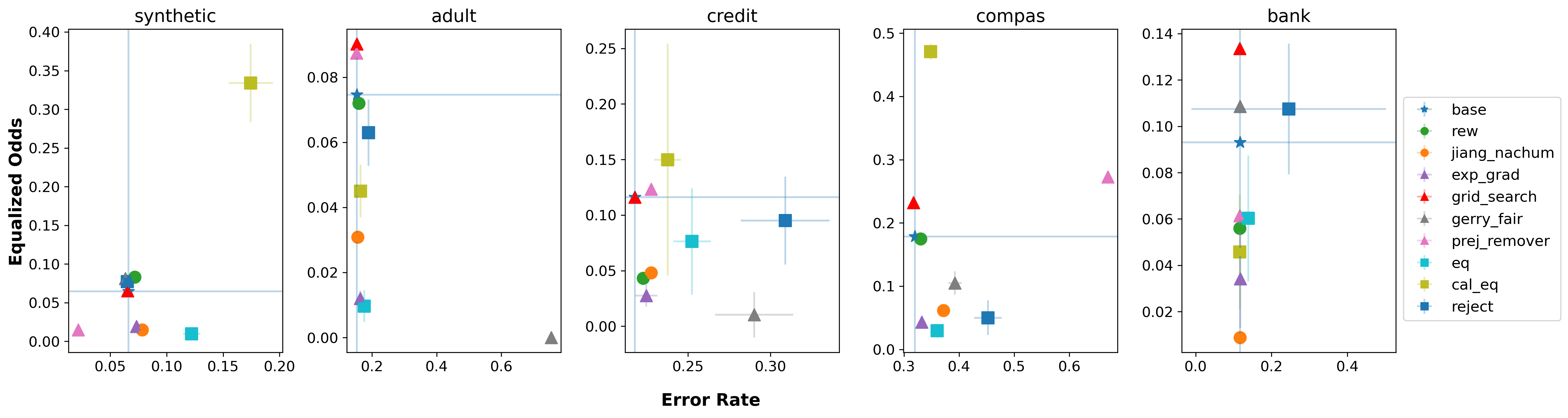}
     \caption{Error Rate-EODDS values for various classifiers on all datasets, with no explicit under-representation and label bias. The blue horizontal and vertical lines denote the error rate and EODDS of a base LR model without any fairness constraints.}
     \label{fig:original_split_results_lr_eodds} 
\end{figure}

We first plot the unfairness (Equalized Odds) and error rates of various fair classifiers on the original training splits. Figure \ref{fig:original_split_results_lr_eodds} shows the results of various fair classifiers with a logistic regression model on original dataset splits without any explicitly injected under-representation or label bias. While most fair classifiers exhibit an expected behavior of trading off some accuracy to minimize unfairness, some classifiers perform worse in mitigating unfairness than the unfair LR model. We attribute this reason to the strong effects of different kinds of data preprocessing on the performance of fair classifiers, as noted in previous works \cite{friedler2019comparative, biswas2021fair}. Other results with SVM as a base classifier, Equal Opportunity Difference, and Statistical Parity Difference are given in Appendix \ref{appndx:original_others}.

We can also inspect Figure \ref{fig:original_split_results_lr_eodds} and other figures later in the paper by looking at the four quadrants formed by the blue horizontal and vertical lines corresponding to the unfairness and error rate of the base LR model. Following a Cartesian coordinate system, most fair classifiers, when trained on the original training set without explicit under-representation or label bias, lie in the fourth quadrant, as they trade-off some accuracy to mitigate unfairness. Any method lying in the first quadrant is poor, as it is worse than the base LR model in terms of both error rate and unfairness. Finally, any method lying in the third quadrant is excellent as it improves over both error rate and unfairness compared to the LR model.

\begin{figure}
     \begin{adjustbox}{minipage=\textwidth}
     \begin{subfigure}{0.3\textwidth}
         \includegraphics[width=\textwidth]{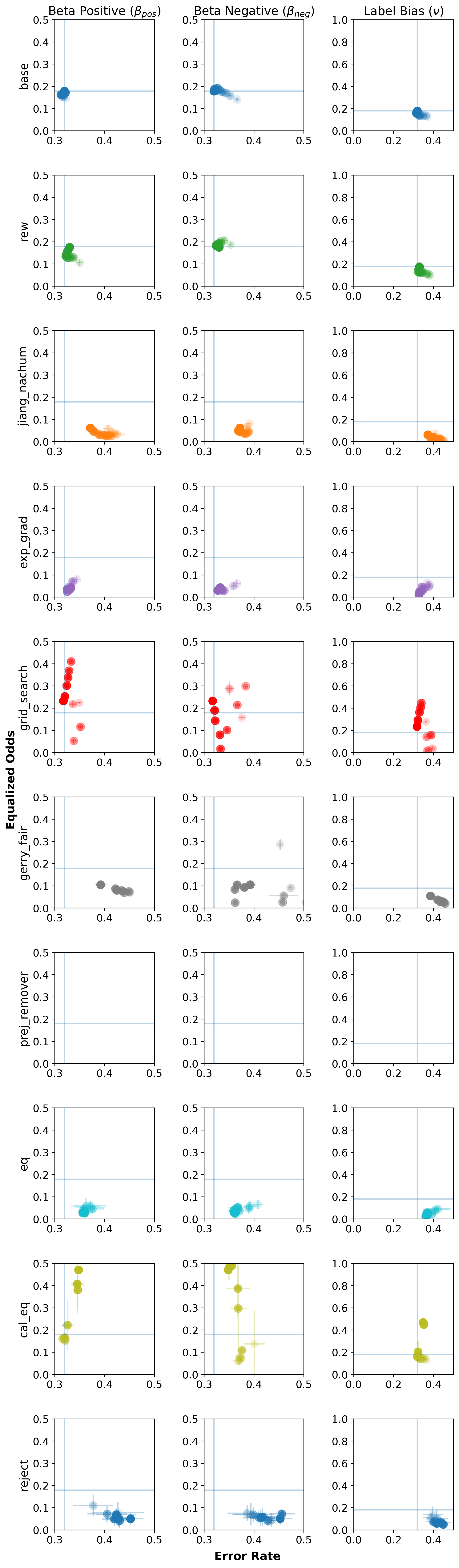}
         \caption{}
     \end{subfigure}
     \begin{subfigure}{0.3\textwidth}
         \includegraphics[width=\textwidth]{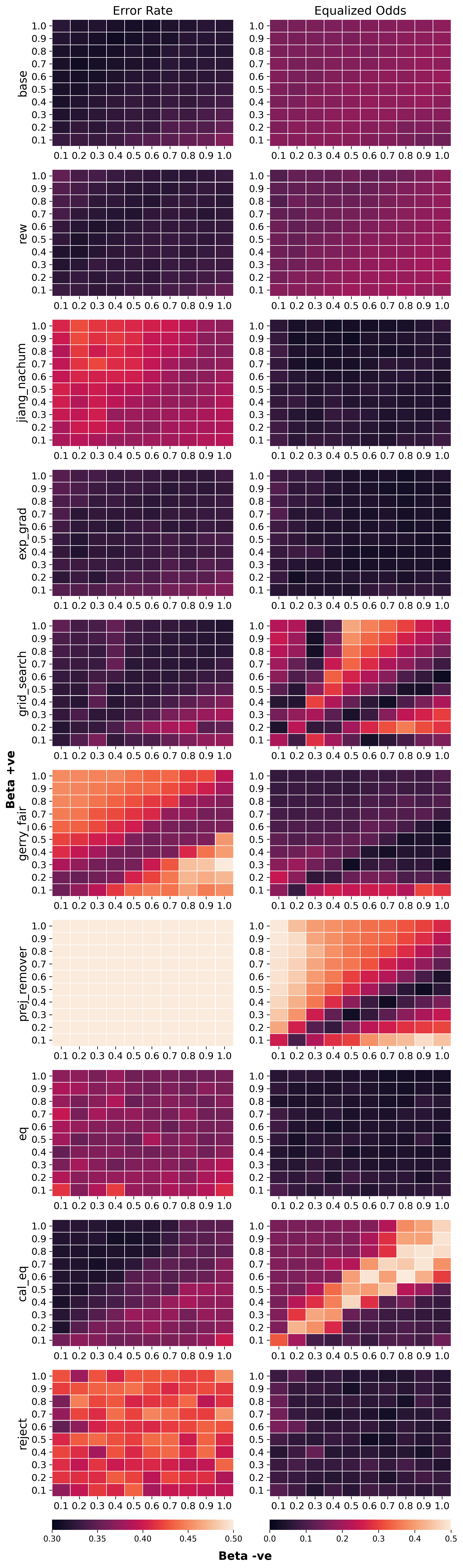}
         \caption{}
     \end{subfigure}
     \begin{subtable}{0.24\textwidth}
            \renewcommand{\arraystretch}{1.33}%
            \centering
            {\notsotiny %
             \begin{tabular}[b]
             {|p{0.7cm}|p{0.3cm}|p{0.75cm}|p{0.75cm}|p{0.75cm}|}
             \hline
              Algo. & & Err. & SPD & EOD \\ \hline
              
              \multirow{2}{*}{base} & lr & \textbf{0.324} $\pm$ \textbf{0.011} & 0.206 $\pm$ \textbf{0.011} & 0.126 $\pm$ 0.024 \\
              \cline{2-5}
               & svm & \textbf{0.338} $\pm$ \textbf{0.013} & 0.186 $\pm$ \textbf{0.014} & 0.106 $\pm$ 0.028 \\
               \hline

               \multirow{2}{*}{rew} & lr & \textbf{0.327} $\pm$ \textbf{0.006} & 0.198 $\pm$ 0.026 & 0.123 $\pm$ 0.037 \\
              \cline{2-5}
               & svm & \textbf{0.343} $\pm$ \textbf{0.008} & 0.177 $\pm$ 0.027 & 0.101 $\pm$ 0.04 \\
               \hline

               \multirow{2}{*}{\parbox{0.9cm}{\centering jiang nachum \hspace{0.1cm}}} & lr & 0.39 $\pm$ 0.014 & \textbf{0.066} $\pm$ \textbf{0.017} & \textbf{0.018} $\pm$ \textbf{0.009} \\
              \cline{2-5}
               & svm & 0.389 $\pm$ \textbf{0.011} & \textbf{0.068} $\pm$ \textbf{0.018} & \textbf{0.016} $\pm$ \textbf{0.006} \\
               \hline

               \multirow{2}{*}{\parbox{0.9cm}{\centering exp \hspace{0.5cm} grad}} & lr & \textbf{0.333} $\pm$ \textbf{0.009} & \textbf{0.07} $\pm$ 0.025 & \textbf{0.029} $\pm$ \textbf{0.022} \\
              \cline{2-5}
               & svm & 0.354 $\pm$ 0.017 & \textbf{0.057} $\pm$ 0.02 & \textbf{0.029} $\pm$ \textbf{0.021} \\
               \hline

               \multirow{2}{*}{\parbox{0.9cm}{\centering grid search}} & lr & 0.336 $\pm$ 0.016 & 0.194 $\pm$ 0.107 & 0.163 $\pm$ 0.096 \\
              \cline{2-5}
               & svm & 0.358 $\pm$ 0.025 & 0.163 $\pm$ 0.092 & 0.13 $\pm$ 0.089 \\
               \hline

               \multirow{2}{*}{\parbox{0.9cm}{\centering gerry fair}} & lr & 0.41 $\pm$ 0.037 & 0.113 $\pm$ 0.057 & 0.097 $\pm$ 0.082 \\
              \cline{2-5}
               & svm & - & - & - \\
               \hline

               \multirow{2}{*}{\parbox{0.9cm}{\centering prej remover \hspace{0.1cm}}} & lr & 0.654 $\pm$ 0.024 & 0.303 $\pm$ 0.137 & 0.246 $\pm$ 0.118 \\
              \cline{2-5}
               & svm & - & - & - \\
               \hline

               \multirow{2}{*}{eq} & lr & 0.37 $\pm$ 0.012 & \textbf{0.05} $\pm$ \textbf{0.016} & 0.044 $\pm$ \textbf{0.021} \\
              \cline{2-5}
               & svm & 0.381 $\pm$ 0.014 & \textbf{0.046} $\pm$ \textbf{0.018} & 0.043 $\pm$ \textbf{0.022} \\
               \hline

               \multirow{2}{*}{cal\_eq} & lr & 0.343 $\pm$ 0.022 & 0.251 $\pm$ 0.141 & 0.171 $\pm$ 0.092 \\
              \cline{2-5}
               & svm & \textbf{0.353} $\pm$ 0.022 & 0.216 $\pm$ 0.114 & 0.144 $\pm$ 0.083 \\
               \hline

               \multirow{2}{*}{reject} & lr & 0.415 $\pm$ 0.019 & 0.079 $\pm$ 0.027 & \textbf{0.038} $\pm$ 0.019 \\
              \cline{2-5}
               & svm & 0.413 $\pm$ 0.021 & 0.081 $\pm$ 0.026 & \textbf{0.041} $\pm$ 0.018 \\
               \hline
               \multicolumn{5}{c}{}
            \end{tabular}}%
         \caption{}
     \end{subtable}

    \caption{Stability analysis of various fair classifiers on the Compas dataset (Lighter the shade, more the under-representation and label bias): (a) The spread of error rates and Equalized Odds(EODDS) of various fair classifiers as we vary $\beta_{pos}$, $\beta_{neg}$ and label bias separately. The vertical and horizontal reference blue lines denote the performance of an unfair logistic regression model on the original dataset without any under-representation or label bias. (b) Heatmap for Error Rate and EODDS across all different settings of $\beta_{pos}$ and $\beta_{neg}$. Darker values denote lower error rates and unfairness. Uniform values across the grid indicate stability to different $\beta_{pos}$ and $\beta_{neg}$. (c) Mean error rates, Statistical Parity Difference (SPD), and Equal Opportunity Difference (EOD) across all $\beta_{pos}$ and $\beta_{neg}$ settings for both kinds of base classifiers (SVM and LR).}
    \label{fig:under_rep_label_bias_compas}
    \end{adjustbox}
\end{figure}

Figure \ref{fig:original_split_results_lr_eodds} shows how various fair classifiers perform without explicit under-representation and label bias. In this figure, we try to summarize all the available pieces of results for a given dataset in various ways to get a holistic picture. We show the stability of error rates and Equalized Odds unfairness and look at the average stability of the other two metrics: Equal Opportunity and Statistical Parity Difference, since many times in practice, we care about the performance of methods against a fixed set of metrics. We now look at the individual effects of both under-representation factors ($\beta_{pos}$, $\beta_{neg}$ and label bias ($\nu$)), with the LR model as a reference, denoted by the vertical and the horizontal error rate and Equal Opportunity Difference (EOD) lines respectively. Figure \ref{fig:under_rep_label_bias_compas}(a) shows those results for the Compas dataset. The results for the Adult, Synthetic, Bank Marketing, and Credit datasets are given in Appendix \ref{appndx:stability_others}. We first observe that many fair classifiers exhibit a large variance in the unfairness-error rate plane and often become unfair as the bias factors increase. This is also indicated by how the fair classifiers jump from quadrant four to quadrant one, as $\beta_{pos}$, $\beta_{neg}$, or $\nu$ increase. In fact, for the `prej\_remover' classifier, the results lie outside the plot limits, indicating its instability. Even when its average SPD is measured in Figure \ref{fig:under_rep_label_bias_compas}(c) (`prej\_remover' is implicitly optimized for SPD), we observe that it does not perform well. 

However, `rew', `jiang\_nachum', and `exp\_grad' fair classifiers remain stable across all three kinds of biases for both error rate and EOD and show a minimal spread across both the error rate and unfairness (EOD) axes. They tend to remain in the same quadrant with varying bias factors. This remains true on other datasets as well (as shown in Appendix \ref{appndx:stability_others}). Furthermore, it's interesting to note that a simple logistic regression model tends to stay relatively stable in the presence of under-representation and label biases, and on many occasions, more fair than many fair classifiers in our analysis.

We can also look at the combined effect of $\beta_{pos}$ and $\beta_{neg}$ in Figure \ref{fig:under_rep_label_bias_compas}(b) with the help of heatmaps. Each cell in the heatmap denotes a possible $\beta_{pos}, \beta_{neg}$ setting, thus covering all the $100$ possible settings. Darker color values in cells denote lower error rates and unfairness. For this analysis, to examine stability with heatmaps, we look for uniformity of colors in the entire grid, which means that the error rate and unfairness values do not change with different $\beta_{pos}, \beta_{neg}$ values. 

Figure \ref{fig:under_rep_label_bias_compas}(b) again confirm the stability of `rew', `jiang\_nachum' and `exp\_grad' fair classifiers even when they are trained on datasets with combined $\beta_{pos}$ and $\beta_{neg}$ biases. However, `jiang\_nachum' and `exp\_grad' emerge as stable and better classifiers in terms of their unfairness values than the `rew' classifier because they have darker uniform colors across their entire error rate and unfairness grid. The `eq' classifier also remains stable for unfairness, except on the Credit dataset (shown in Appendix \ref{appndx:stability_others}). Amongst the classifiers optimized for Equalized Odds (`jiang\_nachum', `exp\_grad', `eq' and `cal\_eq'), `jiang\_nachum' and `exp\_grad' emerge out to be most stable ones in terms of error rate and unfairness. 
 
To quantitatively summarize the findings from the heatmap plots and to show this statistic with other choices of base classifiers (LR and SVM) and unfairness metrics (SPD and EOD), we look at the average performance across $100$ different settings of $\beta_{pos}$ and $ \beta_{neg}$ in Figure \ref{fig:under_rep_label_bias_compas}(c) for the Compas dataset. An indication of stability with these results is whether the mean error rate and unfairness across $100$ different under-representation settings stay low and whether this mean performance has a low standard deviation. Because we run the whole experimental setup $5$ times with different random seeds for reproducibility, the values represent the mean value of the mean across $5$ runs for each $\beta_{pos}, \beta_{neg}$ setting, and its standard deviation over the $100$ settings. We embolden the top-$3$ mean and top-$3$ standard deviation values for each column, where the top-$3$ means the lowest $3$ error rate and the unfairness metrics (SPD and EOD).

For all datasets, the observations across all emboldened values corroborate our findings from the heatmap plots. For both unfairness metrics (SPD and EOD), `exp\_grad' and `jiang\_nachum' often appear in the top-$3$ mean performance across all settings, whereas `rew', `jiang\_nachum' and `exp\_grad' often appear in the top-$3$ smallest standard deviation values, indicating their stability. Furthermore, `rew' and `jiang\_nachum' often appear in the top-$3$ mean and standard deviation list for error rate. `rew' often appears in the top-$3$ mean performance for Statistical Parity Difference (SPD). As expected, the unfair base classifier often appears in the top-$3$ mean error rate list since its objective is always to minimize error rates without regard for unfairness. Finally, when looking over all the datasets' results, we find that `exp\_grad' and `jiang\_nachum' often emerge to be the most stable classifiers for other metrics like EOD and SPD, despite not being trained to minimize those metrics, further solidifying the evidence for their stable behavior. 

Another interesting observation across the results of all datasets is that the deterministic version of the `exp\_grad' fair algorithm, the 'grid\_search' classifier \cite{agarwal2018reductions}, is very unstable compared to the 'exp\_grad' classifier, which is the most stable amongst the lot, indicating that in the face of data bias, randomness might help. Other results, such as Figure \ref{fig:under_rep_label_bias_compas} (a,b) with SPD as the unfairness metric, cannot be included in the manuscript. 

Finally, it is worth noting that the stability of `rew' and `exp\_grad' has also been reported in another benchmarking study about testing various fair classifiers in the presence of varying sensitive attribute noise \cite{ghosh2023fair}. This indicates towards a potentially stable performance from reweighing-based approaches in the presence of varying kinds of data biases and noises in the training data.

\section{Stability Theorems for Reweighing Classifier} \label{sec:theorems}
The following theorems give a theoretical justification for the stability of test accuracy when the reweighing classifier \cite{kamiran2012data} is trained even on data injected with extreme under-representation such as $\beta_{\text{pos}} \rightarrow 0$ or $\beta_{\text{neg}} \rightarrow 0$.

\begin{theorem} \label{thm:sandwich}
Let $D$ be the original data distribution and $\beta \in (0, 1]$. Let $D_{\beta}$ be the biased data distribution after injecting under-representation bias in $D$ using $\beta_{\text{pos}} = \beta$ (or similarly $\beta_{\text{neg}} = \beta$). Let $f$ be any binary classifier and $L(f)$ be any non-negative loss function on the prediction of $f$ and the true label. Let $L'$ denote the reweighed loss optimized by the reweighing classifier \cite{kamiran2012data}. Then
\[
\frac{\alpha^{2}}{4}~ \mathbb{E}_{D}[L(f)] \leq \mathbb{E}_{D_{\beta}}[L'(f)] \leq \frac{4}{\alpha}~ \mathbb{E}_{D}[L(f)],
\]
where $\alpha = \frac{\min_{ij} \operatorname{P}\left(Y=i, S=j\right)}{\max_{ij} \operatorname{P}\left(Y=i, S=j\right)}$ denotes the subgroup imbalance in the original distribution $D$.
\end{theorem}

The proof for this theorem is given in Appendix \ref{appndx:proofs}. The proof involves bounding subgroup and group/label probabilities using $\alpha$, and with a reasonable assumption that  $X'| Y'=y, S'=s$ in $D_{\beta} \sim X| Y=y, S=s$ in $D$. The main take-away from Theorem \ref{thm:sandwich} is the following: when we reweigh a given non-negative loss function using the Reweighing scheme from Kamiran \& Calders \cite{kamiran2012data} on samples from $D_{\beta}$, we always lie in some constant `radius' of the expected loss on the true distribution $D$ for any classifier $f$. This does not tell us anything about the expected loss on $D$ itself, but rather, about the ability of the reweighed loss $L'$ to recover from any arbitrary under-representation bias $\beta$. However, the constant factors depend on $\alpha$, the worst case imbalance between the subgroups for a given distribution. 

\begin{remark}
    Whenever $\mathbb{E}_{D}[L(f)]$ is small, Theorem \ref{thm:sandwich} says that the expected reweighed loss on the biased distribution $\mathbb{E}_{D_{\beta}}[L'(f)]$ will also be small, for any classifier $f$. This means that if we are given a non-negative loss function which also subsumes a given fairness constraint (let's call it $L_{fair}$), then reweighing with $L_{fair}$ can give us low unfairness and stable classifiers.
\end{remark}

Using Theorem \ref{thm:sandwich}, we prove the following bound on the expected loss of the reweighing classifier that is obtained by empirical minimization of the reweighed loss on the biased distribution $D_{\beta}$ but tested on the original distribution $D$.

\begin{theorem} \label{thm:loss}
For any $\beta \in (0, 1]$, let $\hat{g}_{\beta}$ the classifier obtained by empirical minimization of the reweighed loss $L'$ on $N$ samples from the biased distribution $D_{\beta}$ defined as in Theorem \ref{thm:sandwich}. Then, with probability at least $1 - \delta$, we have
\[
\mathbb{E}_{D}[L(\hat{g_{\beta}})] \leq \frac{16}{\alpha^{3} \beta}~ \sqrt{\frac{\ln{(2/\delta)}}{2N}} + \frac{16}{\alpha^{3}}~ \mathbb{E}_{D}[L(f^{*})],
\]
where $f^{*}$ is the Bayes optimal classifier on the original distribution $D$, $\alpha$ is the subgroup imbalance as in Theorem \ref{thm:sandwich}.
\end{theorem}

The proof for this theorem is given in Appendix \ref{appndx:proofs}. It uses a generalized version of the Hoeffding's inequality \cite{hoeffding1963probability}, proof of Theorem $2$ of Zhu et al. \cite{zhu2021rich} and Theorem ~\ref{thm:sandwich}. Note that Theorem \ref{thm:loss} implies that given any $\epsilon > 0$ and $\delta > 0$, we can get the guarantee
\[
\mathbb{E}_{D}[L(\hat{g}_{\beta})] \leq \frac{16}{\alpha^{3}}~ \mathbb{E}_{D}[L(f^{*})] + \epsilon,
\]
with probability at least $1-\delta$, by simply choosing the number of samples $N$ from $D_{\beta}$ in the empirical reweighed loss minimization to be large enough so that
\[
N \geq \frac{128 \log(2/\delta)}{\alpha^{6} \beta^{2} \epsilon^{2}}.
\]

It is important to note that Theorems \ref{thm:sandwich} and \ref{thm:loss} hold for data without any injected bias, i.e., $\beta_{\text{pos}} = \beta_{\text{neg}} = 1$, where such theoretical guarantees for the reweighing classifier \cite{kamiran2012data} were not known earlier.

\section{Discussion} \label{sec:discussion}
In this section, we discuss where our results stand in the context of related works on distribution shifts and data bias in real-world AI/ML models and make a few practical recommendations based on our work.

\subsection{Under-Representation and Distribution Shifts}
Let $P_{\text{train}}$ and $P_{\text{test}}$ define the probabilities for training and testing distributions, respectively, with a random data point denoted by $(X, S, Y)$ with features $X$, sensitive attribute $S$, and class label $Y$. Let $Y=1$ be the favorable label and $S=0$ be the underprivileged group. A covariate shift in fair classification is defined as $P_{\text{train}}(Y|X,S) = P_{\text{test}}(Y|X,S)$ \cite{rezaei2020robust}.
Singh et al. \cite{singh2021fairness} look at fairness and covariate shift from a causal perspective using the notion of a separating set of features that induce a covariate shift. Coston et al. \cite{coston2019fair} look at covariate shift and domain adaptation when sensitive attributes are not available at source (train) or target (test).

Distribution shifts can also be induced via class labels. Dai et al. \cite{dailabel} present a more general model for label bias, building upon the work of Blum et al. \cite{blum2019recovering}. They present a model for label shift where $P_{\text{train}}(Y) \neq P_{\text{test}}(Y)$, but $P_{\text{train}}(X|Y,S) = P_{\text{test}}(X|Y,S)$. Biswas et al. \cite{biswas2021ensuring} study model shifts in prior probability, where $P_{\text{train}}(Y=1|S) \neq P_{\text{test}}(Y=1|S)$ but $P_{\text{train}}(X|Y=1,S) = P_{\text{test}}(X|Y=1,S)$. Our work is also different from sample selection bias \cite{cortes2008sample} (selection of training samples at random vs. sensitive group dependent under-sampling and label flipping), but may have some connections to fair sample selection bias \cite{du2021fair}, where the objective is to train only on selected samples($\mathscr{S}=1$), but generalize well for fairness and accuracy on the overall population (unselected samples $\mathscr{S}=0$ and $\mathscr{S}=1$). An interesting question one can ask in this regard is to design fair classifiers which are stable against a range of malicious sample selection biases. 


The under-representation model used in our paper comes from Blum et al. \cite{blum2019recovering} and can be thought of as the following distribution shift: $P_{\text{train}}(Y=0,S=0) = \beta_{\text{neg}} \cdot P_{\text{test}}(Y=0,S=0) \text{ and } P_{\text{train}}(Y=1,S=0) = \beta_{\text{pos}} \cdot P_{\text{test}}(Y=1,S=0)$, while the other two subgroups ($Y=0,S=1$) and ($Y=1,S=1$) are left untouched. Overall, the distribution of $(X,S)$ remains unchanged, while the distribution of $Y$ changes only for the underprivileged group $S=0$. The above distribution shift is different from covariate and label shifts in that it affects both the joint marginal $P(X, S)$ and $P(X|Y, A)$. 

In a broader sense beyond the mathematical definition used in our work, under-representation of a demographic and implicit bias in the labels are known problems observed in real-world data. Shankar et al. \cite{shankar2017no} highlight geographic over-representation issues in public image datasets and how that can harm use cases concerning prediction tasks involving developing countries. Buolamwini et al. \cite{buolamwini2018gender} highlight a similar under-representation issue with gender classification systems trained on over-represented lighter-skinned individuals. Wilson et al. \cite{wilson2019predictive} highlight similar disparities for pedestrian detection tasks. Biased labels in the training data have also been observed in the context of implicit bias in the literature \cite{greenwald2006implicit, kleinberg2018selection}. 

\subsection{Practical Recommendations}
We show that the performance guarantees of commonly used fair classifiers can be highly unstable when their training data has under-representation and label bias. Our experimental setup serves as a template of a dashboard to check the vulnerability of fair classifiers to data bias by injecting bias using simple models similar to Blum \& Stangl \cite{blum2019recovering}.
    

Our work motivates the importance of creating test suites for robustness to data bias and incorporating them with modern fairness toolkits such as AIF-360 \cite{aif360-oct-2018}. 
Our results complement previous studies on the stability across different train-test splits \cite{friedler2019comparative} or different data pre-processing strategies \cite{biswas2021fair}. We experiment with a range of $\beta_{pos}, \beta_{neg}$ and $\nu$ factors. In practice, this can be done with a user-supplied range of under-representation and label bias relevant for specific use cases. The ability to measure the stability of chosen fairness metrics across dynamically specified bias factors can be a great first step towards safe deployment of fair classifiers, similar to recent works like Shifty \cite{giguere2022fairness}.

Our results show that some classifiers like `jiang\_nachum' \cite{jiang2020identifying} and `exp\_grad' \cite{agarwal2018reductions} can remain relatively unscathed in the presence of varying under-representation and label biases. The stability is maintained even when they are trained and tested on mismatching fairness metrics. This motivates using such reweighing methodologies when the level of data biases our model might see is unknown.

Motivated by the same setup of Blum \& Stangl \cite{blum2019recovering}, Akpinar et al. \cite{akpinar2022sandbox} also propose a toolkit and dashboard to stress test fair algorithms under a user-controlled synthetic setup, which allows the user to control various aspects of the simulation. A handy addition to existing fairness toolkits such as AIF360 \cite{aif360-oct-2018}, and Fairlearn \cite{bird2020fairlearn} can be to incorporate input distribution stability routines from our work and Akpinar et al. \cite{akpinar2022sandbox}, along with some comparative analysis routines from Friedler et al. \cite{friedler2019comparative} to create a practical checklist for any fairness algorithm. Finally, our experimental setup can provide an additional tool for checking the robustness of models to data bias, along with the existing robustness test suites, e.g.,  CheckList \cite{ribeiro2020beyond}.

\section{Conclusion} \label{sec:conclusion}
We show that many state-of-the-art fair classifiers exhibit a large variance in their performance when their training data is injected with under-representation and label bias. We propose an experimental setup to compare different fair classifiers under data bias. We show how the Bayes Optimal unfair and fair classifiers react with under-representation bias and propose a reweighing scheme to recover the true fair Bayes Optimal classifiers while optimizing over biased distributions with a known bias. We give a theoretical bound on the accuracy of the reweighing classifier \cite{kamiran2012data} that holds even under extreme data bias. Finally, we discuss our work in the broader context of data bias literature and make practical recommendations.

A limitation of our work is that we use a simple model for injecting under-representation and label bias, with which one cannot hope to address the root cause behind different kinds of biases in real-world data. Thinking about the causal origin of data bias instead may allow one to model more types of biases and obtain better fixes for them under reasonable assumptions. Theoretically explaining the stability of the `exp\_grad' \cite{agarwal2018reductions}, and `jiang\_nachum' \cite{jiang2020identifying} fair classifiers with under-representation bias is also an interesting future work.

\bibliographystyle{plain}
\bibliography{mybibfile}

\appendix

\section{Detailed Related Work} \label{appndx:related_work}

\subsection{Prior Work on Distribution Shifts} \label{appndx:related_work_dist_shift}

The robustness of ML models when there is a mismatch between the training and test distributions has been studied under various theoretical models of distribution shifts, e.g., covariate shift and label shift. Recent works have studied fair classification subject to these distribution shifts and proposed solutions under reasonable assumptions on the data distribution \cite{coston2019fair, schumann2019transfer, blum2019recovering, rezaei2020robust, singh2021fairness, dailabel, biswas2021ensuring, du2021fair, maity2021does, giguere2022fairness}. 

Coston et al. \cite{coston2019fair} study the problem of fair transfer learning with covariate shift under two conditions: availability of protected attributes in the source or target data, and propose to solve these problems by weighted classification and using loss functions with twice differentiable score disparity terms. Schumann et al. ~\cite{schumann2019transfer} present theoretical guarantees on different scenarios of fair transfer learning for equalized odds and equal opportunity through a domain adaptation framework. Rezaei et al. ~\cite{rezaei2020robust} propose an adversarial framework to learn fair predictors in the presence of covariate shift by formulating a game between choosing fair minimizing and maximizing target conditional label distributions, formulating a convex optimization problem which they ultimately solve using batch gradient descent. Singh et al. ~\cite{singh2021fairness} proposes to learn stable models with respect to fairness and accuracy in the presence of shifts in data distribution using a causal graph modeling the shift, using the framework of causal domain adaptation, deriving solutions that are worst-case optimal for some kinds of covariate shift. Dai et al. ~\cite{dailabel} talk about the impact of label bias and label shift on standard fair classification algorithms. They study label bias and label shift through a common framework where they assume the availability of a trained fair classifier on true labels and analyze its performance at test time when either the true label distribution changes or when the new test time labels are biased. Biswas et al. ~\cite{biswas2021ensuring} focus on addressing the effects of prior probability shifts on fairness and propose an algorithm and a metric called Prevalence Difference, which, when minimized, ensures Proportional Equality. 

Finally, Du et al. ~\cite{du2021fair} propose a framework to realize classifiers robust to sample selection bias using reweighing and minimax robust estimation. The selection of samples is indicated by $\mathscr{S}$. The aim is to train using a part of the distribution where $\mathscr{S}=1$ but generalize well for fairness and accuracy on the overall population (both $\mathscr{S}=0$ and $\mathscr{S}=1$). Du et al. \cite{du2021fair} show that robust and fair classifiers on the overall population can be obtained by solving a reweighed optimization problem. Giguere et al. \cite{giguere2022fairness} consider demographic shift where a certain group can become more or less probable in the test distribution than the training distribution. They output a classifier that retains fairness guarantees for given user-specified bounds on these probabilities. Zhu et al. \cite{zhu2023consistent} study the fair sample selection bias problem and propose the CRAB framework, which constructs a causal graph based on the background knowledge about the data collection process and external data sources.

We tackle a related problem to the above works, where we only look at the performance of fair classifiers in varying amounts of under-representation bias (or subgroups imbalance) and label bias, which is a very straightforward case of distribution shift. 

\subsection{Prior Work on Comparative Analysis of Fair Methods} \label{appndx:related_work_compare}

We compare various fair classifiers in the presence of under-representation and label bias. Our work complements the surveys in fair machine learning that compare different methods either conceptually \cite{zliobaite2015survey, mehrabi2021survey, pessach2022review, feng2022fair} or empirically, on important aspects like the choice of data processing, splits, data errors, and data pre-processing choices \cite{friedler2019comparative, biswas2021ensuring, qian2021my, islam2022through}. Friedler et al. \cite{friedler2019comparative} compare some fair classification methods in terms of how the data is prepared and preprocessed for training, how different fairness metrics correlate with each other, stability with respect to slight changes in the training data, and the usage of multiple sensitive attributes. Islam et al. \cite{islam2022through} conceptually compare various fairness metrics and further use that to empirically compare $13$ fair classification approaches on their fairness-accuracy tradeoffs, scalability with respect to the size of dataset and dimensionality, robustness against training data errors such as swapped and missing feature values, and variability of results with multiple training data partitions. On the other hand, we compare various classifiers with respect to changes in the distribution while keeping most things like base classifiers and data preprocessing the same across methods.

\section{Dataset Information} \label{appndx:dataset_info}

Table \ref{tab:dataset_info} list all information pertaining to the datasets used in our experiments. We choose to drop columns which either introduce a lot of empty values, or represent some other sensitive attribute. The reason to drop the non-primary sensitive attributes is to avoid unintended disparate treatment. The choice of sensitive attributes for each dataset is motivated from previous works in fair ML literature \cite{zafar2017fairness, agarwal2018reductions, friedler2019comparative, jiang2020identifying}.

The synthetic dataset setup from Zeng et al. \cite{zeng2022bayes} consists of a binary label $y \in \{0,1\}$ and a binary sensitive attribute setting $s \in \{0,1\}$. We sample data from subgroup-specific $100$-dimensional Gaussian distributions $X \sim \mathcal{N}(\mu_{sy},\sigma^{2}I_{100})$ where $I_{100}$ denotes a $100 \times 100$ identity matrix and all entries of $\mu_{sy}$ are sampled from $U(0,1)$. We sample $20,000$ training examples and $5000$ test examples.

\setlength{\textfloatsep}{1pt}
\begin{table}[]
    \centering
    \begin{center}
        \begin{tabular}{|p{2cm}|c|p{2cm}|p{3cm}|p{3cm}|} 
             \hline
             \textbf{Dataset} & \textbf{\# Data Pts.} & \textbf{Outcome (Y)} & \textbf{Sensitive Attribute(S)} & \textbf{Dropped Columns} \\
             \hline
             Adult \cite{Dua:2019} & $45k$ & Income(>$50k$, <=$50k$) & Sex (Male, Female) & fnlwgt, education-num, race \\
             \hline
             Bank \cite{moro2014data} & $41k$ & Term Deposit (Yes, No) & Age (<=$25$, >$25$) & poutcome, contact \\
             \hline
             Compas \cite{angwin2016machine} & $6k$ & Recidivate (Yes,No) & Race(African-American, Caucasian) & sex \\
             \hline
             Credit \cite{Dua:2019} & $1k$ & Credit Risk (good, bad) &Sex (Male, Female) & age, foreign\_worker \\
             \hline
             Synthetic \cite{zeng2022bayes} & $20k$ & $0,1$ & $0,1$ & NA \\
             \hline
        \end{tabular}
    \end{center}
    \caption{Details of all the datasets used in our Experiments.}
    \label{tab:dataset_info}
\end{table}

\section{Algorithm to create different Data Bias scenarios} \label{appndx:algorithm}

Algorithm \ref{algo:bias} gives a detailed description of how we inject varying amounts of under-representation and label bias for the considered fair classifiers.

\setlength{\textfloatsep}{1pt}
\begin{algorithm}[t]
\small
\KwIn{A fair classifier, or a base classifier (lr or svm without fairness constraints): $cls$, original training dataset: $data\_train$, test dataset: $data\_test$. All samples in $data\_train$ and $data\_test$ are of the form $(x,y,s)$}
\KwOut{Results for all under-representation and label bias setting for $cls$ on the test set, with multiple runs.}
$all\_runs\_results$ = \{\}

\For{$run \leftarrow 1$ \KwTo $5$}{
    $under\_rep\_results$ = \{\}, $label\_bias\_results$ = \{\}
    
    \For{$\beta_{pos} \in [0.1 .... 1.0]$}{
        \For{$\beta_{neg} \in [0.1 .... 1.0]$}{
            $train\_data$($\beta_{pos}, \beta_{neg}$) $\leftarrow$ Randomly sample $\beta_{pos}$ proportion of ($y=1, s=0$) subpopulation and $\beta_{neg}$ proportion of ($y=0, s=0$) subpopulation from $data\_train$, include other subpopulations same as $data\_train$.
            
            $under\_rep\_results$[($\beta_{pos}, \beta_{neg}$)] $\leftarrow$ result for $cls$ on $train\_data$($\beta_{pos}, \beta_{neg}$)
        }
    }

    \For{$\nu \in [0.0 .... 0.9]$}{
        $train\_data$($\nu$) $\leftarrow$ Randomly flip $\nu$ proportion of ($y=1, s=0$) subpopulation examples to ($y=0, s=0$) from $data\_train$, include other subpopulations same as $data\_train$.
        
        $label\_bias\_results$[$\nu$] $\leftarrow$ result for $cls$ on $train\_data$($\nu$)
    }

    $all\_runs\_results$[$run$] = ($under\_rep\_results$, $label\_bias\_results$)
}

\KwRet{all\_runs\_results}
\caption{Experimental Pipeline for Under-Representation and Label Bias}
\label{algo:bias}
\end{algorithm}

\section{Original Split results for other settings} \label{appndx:original_others}

\begin{figure}
     \centering
     \includegraphics[width=\textwidth]{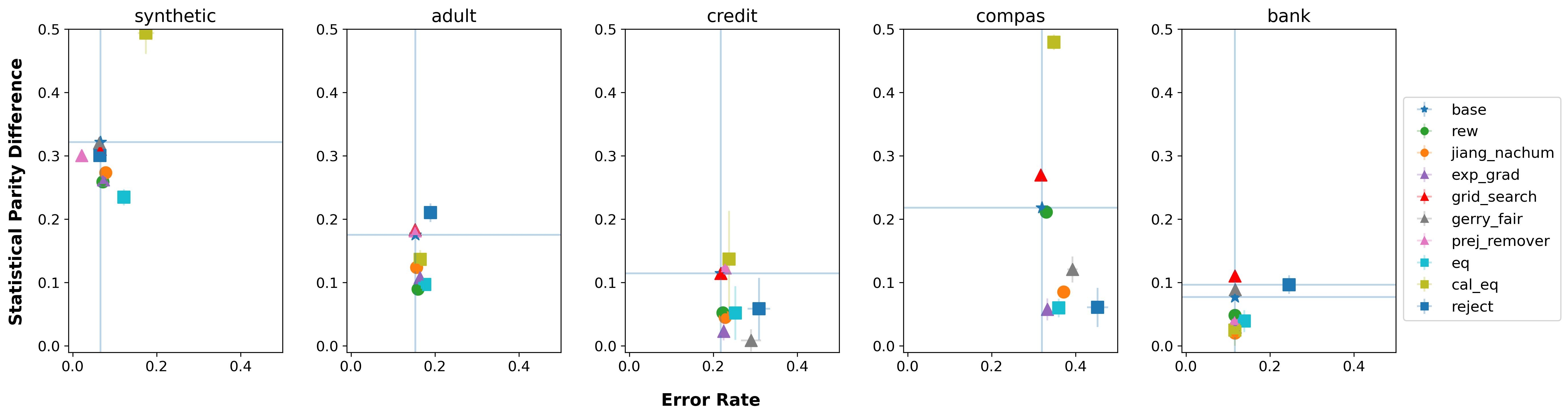}
     \caption{Error Rate-SPD values for various classifiers on all datasets, with no explicit under-representation and label bias. The blue horizontal and vertical lines denote the error rate and SPD of a base LR model without any fairness constraints.}
     \label{fig:original_split_results_lr_dp} 
\end{figure}

\begin{figure}
     \centering
     \includegraphics[width=\textwidth]{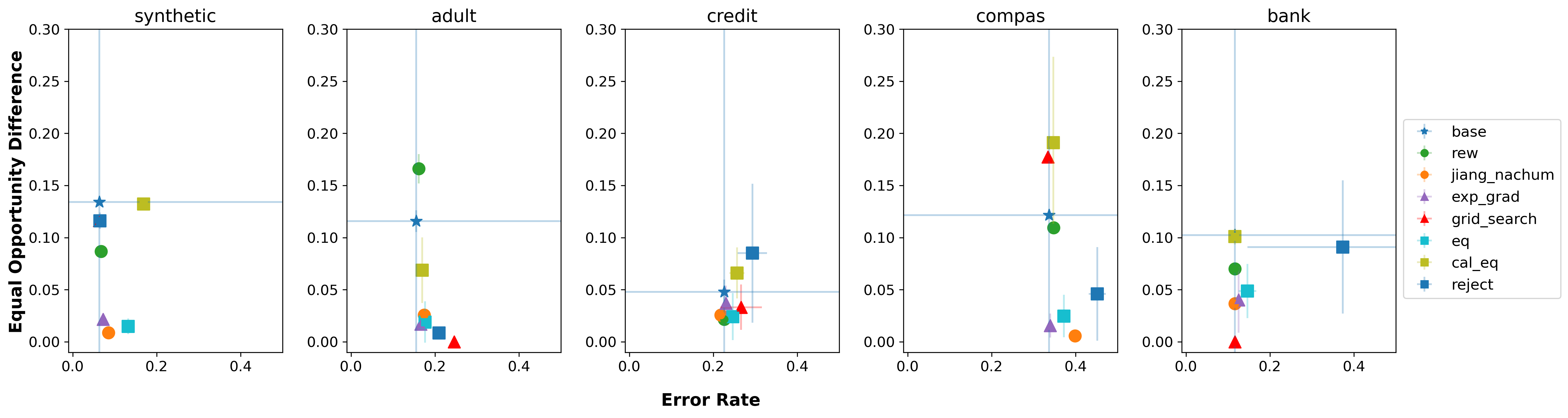}
     \caption{Error Rate-EOD values for various classifiers on all datasets, with no explicit under-representation and label bias. The blue horizontal and vertical lines denote the error rate and EOD of a base SVM model without any fairness constraints.}
     \label{fig:original_split_results_svm_eod} 
\end{figure}

\begin{figure}
     \centering
     \includegraphics[width=\textwidth]{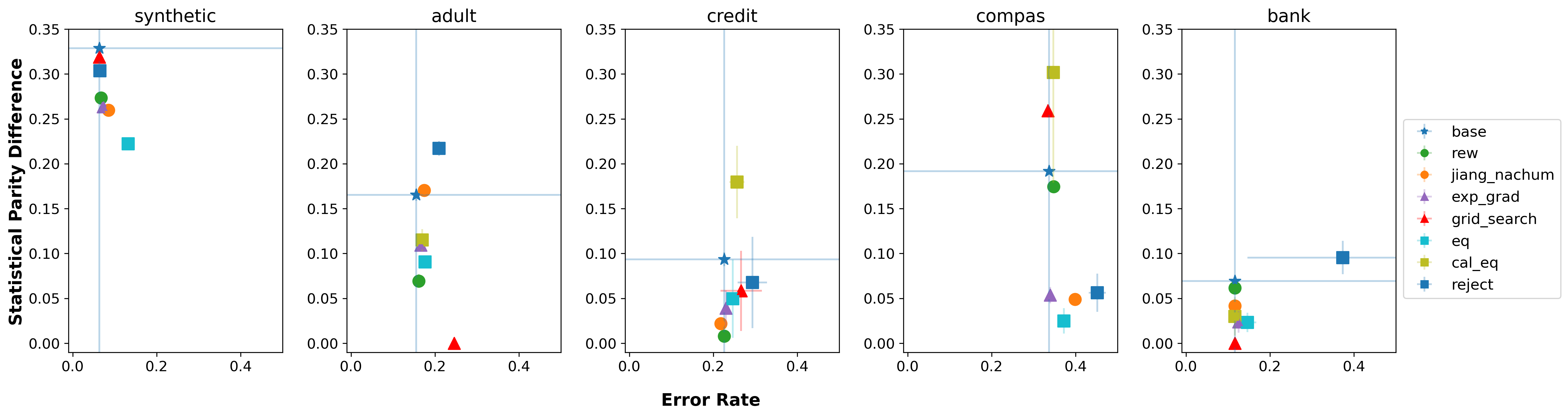}
     \caption{Error Rate-SPD values for various classifiers on all datasets, with no explicit under-representation and label bias. The blue horizontal and vertical lines denote the error rate and SPD of a base SVM model without any fairness constraints.}
     \label{fig:original_split_results_svm_dp} 
\end{figure}

Figures \ref{fig:original_split_results_lr_dp}, \ref{fig:original_split_results_svm_eod} and \ref{fig:original_split_results_svm_dp} show the results of all fair classifiers when we use either a Logistic Regression model or an SVM model, and measure unfairness using Statistical Parity or Equal Opportunity Difference as a metric. Note that `gerry\_fair' and `prej\_remover' are missing from Figures \ref{fig:original_split_results_svm_eod} and \ref{fig:original_split_results_svm_dp} because these two classifiers do not support SVM as a base classifier for their fairness routines. 

\begin{figure}
     \centering
     \includegraphics[width=\textwidth]{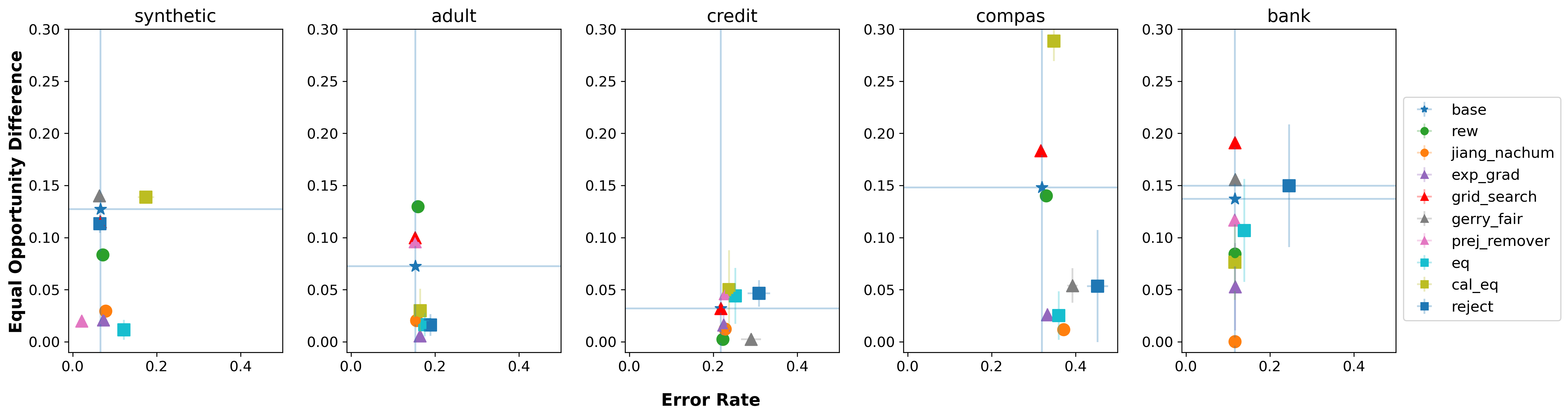}
     \caption{Error Rate-EOD values for various classifiers on all datasets, with no explicit under-representation and label bias. The blue horizontal and vertical lines denote the error rate and EOD of a base LR model without any fairness constraints.}
     \label{fig:original_split_results_lr_eod} 
\end{figure}

\section{Stability Analysis for remaining datasets} \label{appndx:stability_others}
\label{appendix_stability}

\begin{figure}
     \begin{adjustbox}{minipage=\textwidth}
     \begin{subfigure}{0.3\textwidth}
         \includegraphics[width=\textwidth]{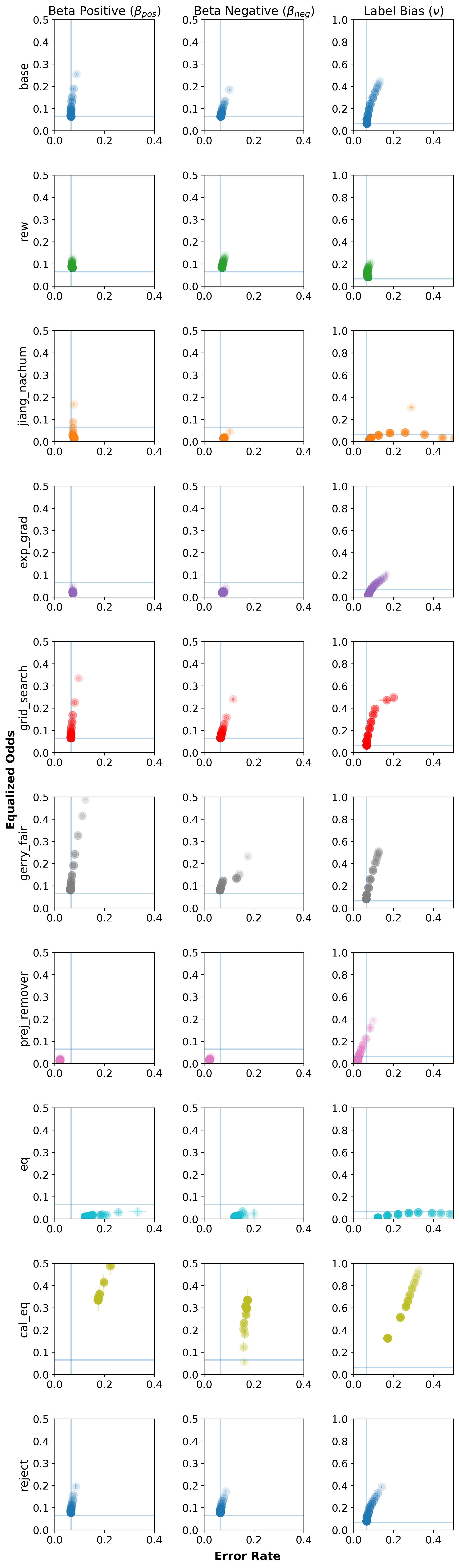}
         \caption{}
     \end{subfigure}
     \begin{subfigure}{0.3\textwidth}
         \includegraphics[width=\textwidth]{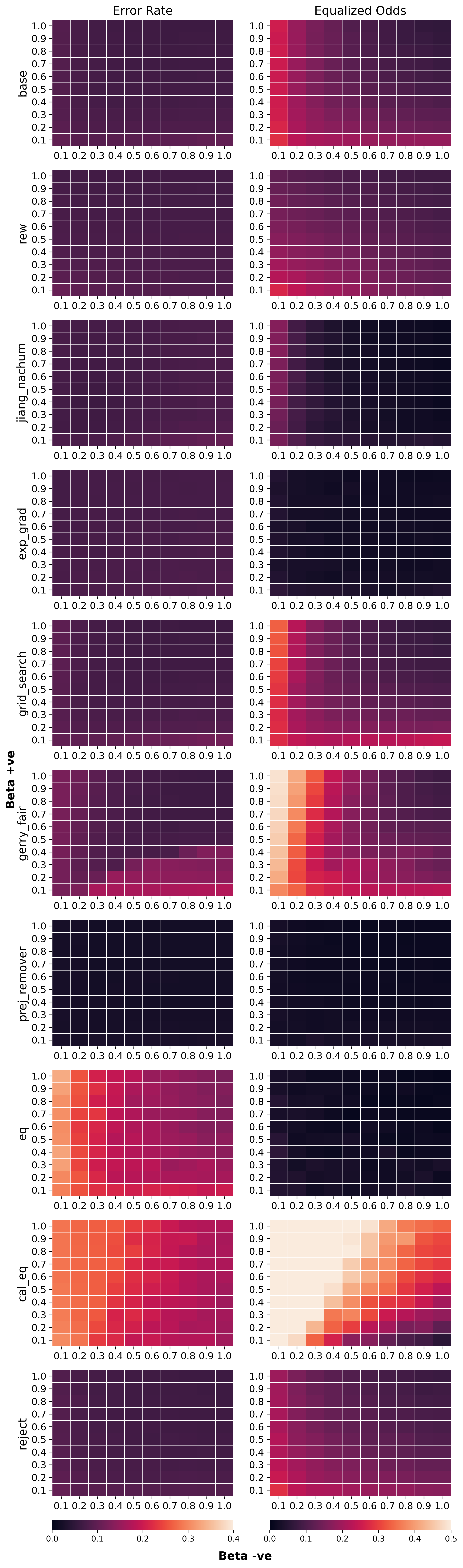}
         \caption{}
     \end{subfigure}
     \begin{subtable}{0.24\textwidth}
            \renewcommand{\arraystretch}{1.35}%
            \centering
            {\notsotiny %
             \begin{tabular}[b]
             {|p{0.7cm}|p{0.3cm}|p{0.75cm}|p{0.75cm}|p{0.75cm}|}
             \hline
              Algo. & & Err. & SPD & EOD \\ \hline
              
              \multirow{2}{*}{base} & lr & 0.075 $\pm$ 0.009 & 0.311 $\pm$ 0.073 & 0.206 $\pm$  0.107 \\
              \cline{2-5}
               & svm & \textbf{0.074} $\pm$ 0.009 & 0.325 $\pm$  0.074 & 0.218 $\pm$ 0.115 \\
               \hline

               \multirow{2}{*}{rew} & lr & 0.077 $\pm$ 0.006 & \textbf{0.241} $\pm$ \textbf{0.018} & 0.122 $\pm$  0.033 \\
              \cline{2-5}
               & svm & \textbf{0.074} $\pm$ \textbf{0.006} & \textbf{0.258} $\pm$ \textbf{0.023} & 0.131 $\pm$ \textbf{0.04} \\
               \hline

               \multirow{2}{*}{\parbox{0.9cm}{\centering jiang nachum \hspace{0.1cm}}} & lr & \textbf{0.075} $\pm$ \textbf{0.006} & 0.291 $\pm$ 0.041 & 0.086 $\pm$ 0.079 \\
              \cline{2-5}
               & svm & 0.083 $\pm$ 0.009 & 0.289 $\pm$ 0.046 & \textbf{0.086} $\pm$ 0.093 \\
               \hline

               \multirow{2}{*}{\parbox{0.9cm}{\centering exp \hspace{0.5cm} grad}} & lr & 0.076 $\pm$ \textbf{0.003} & \textbf{0.259} $\pm$ \textbf{0.009} & \textbf{0.029} $\pm$ \textbf{0.011} \\
              \cline{2-5}
               & svm & 0.079 $\pm$ \textbf{0.005} & \textbf{0.257} $\pm$ \textbf{0.01} & \textbf{0.026} $\pm$ \textbf{0.01} \\
               \hline

               \multirow{2}{*}{\parbox{0.9cm}{\centering grid search}} & lr & 0.078 $\pm$ 0.012 & 0.301 $\pm$ 0.111 & 0.207 $\pm$ 0.149 \\
              \cline{2-5}
               & svm & 0.078 $\pm$ 0.013 & 0.315 $\pm$ 0.119 & 0.22 $\pm$ 0.167 \\
               \hline

               \multirow{2}{*}{\parbox{0.9cm}{\centering gerry fair}} & lr & 0.099 $\pm$ 0.032 & 0.368 $\pm$ 0.144 & 0.354 $\pm$ 0.245 \\
              \cline{2-5}
               & svm & - & - & - \\
               \hline

               \multirow{2}{*}{\parbox{0.9cm}{\centering prej remover \hspace{0.1cm}}} & lr & \textbf{0.022} $\pm$ \textbf{0.001} & 0.296 $\pm$ \textbf{0.009} & \textbf{0.022} $\pm$ \textbf{0.008} \\
              \cline{2-5}
               & svm & - & - & - \\
               \hline

               \multirow{2}{*}{eq} & lr & 0.192 $\pm$ 0.052 & \textbf{0.187} $\pm$ 0.029 & \textbf{0.025} $\pm$ \textbf{0.017} \\
              \cline{2-5}
               & svm & 0.191 $\pm$ 0.052 & \textbf{0.188} $\pm$ \textbf{0.028} & \textbf{0.025} $\pm$ \textbf{0.017} \\
               \hline

               \multirow{2}{*}{cal\_eq} & lr & 0.218 $\pm$ 0.042 & 0.534 $\pm$ 0.132 & 0.219 $\pm$ 0.11 \\
              \cline{2-5}
               & svm & 0.22 $\pm$ 0.042 & 0.535 $\pm$ 0.132 & 0.22 $\pm$ 0.111 \\
               \hline

               \multirow{2}{*}{reject} & lr & \textbf{0.075} $\pm$ 0.008 & 0.278 $\pm$ 0.028 & 0.168 $\pm$ 0.043 \\
              \cline{2-5}
               & svm & \textbf{0.075} $\pm$ \textbf{0.008} & 0.276 $\pm$ 0.028 & 0.167 $\pm$ \textbf{0.046} \\
               \hline
               \multicolumn{5}{c}{}
            \end{tabular}}%
         \caption{}
     \end{subtable}

    \caption{Stability analysis of various fair classifiers on the Synthetic dataset (Lighter the shade, more the under-representation and label bias): (a) The spread of error rates and Equalized Odds (EODDS) of various fair classifiers as we vary $\beta_{pos}$, $\beta_{neg}$ and label bias separately. The vertical and horizontal reference blue lines denote the performance of an unfair logistic regression model on the original dataset without any under-representation or label bias. (b) Heatmap for Error Rate and EODDS across all different settings of $\beta_{pos}$ and $\beta_{neg}$. Darker values denote lower error rates and unfairness. Uniform values across the grid indicate stability to different $\beta_{pos}$ and $\beta_{neg}$. (c) Mean error rates, Statistical Parity Difference (SPD), and Equal Opportunity Difference (EOD) across all $\beta_{pos}$ and $\beta_{neg}$ settings for both kinds of base classifiers (SVM and LR).}
    \label{fig:under_rep_label_bias_synthetic}
    \end{adjustbox}
\end{figure}

\begin{figure}
     \begin{adjustbox}{minipage=\textwidth}
     \begin{subfigure}{0.3\textwidth}
         \includegraphics[width=\textwidth]{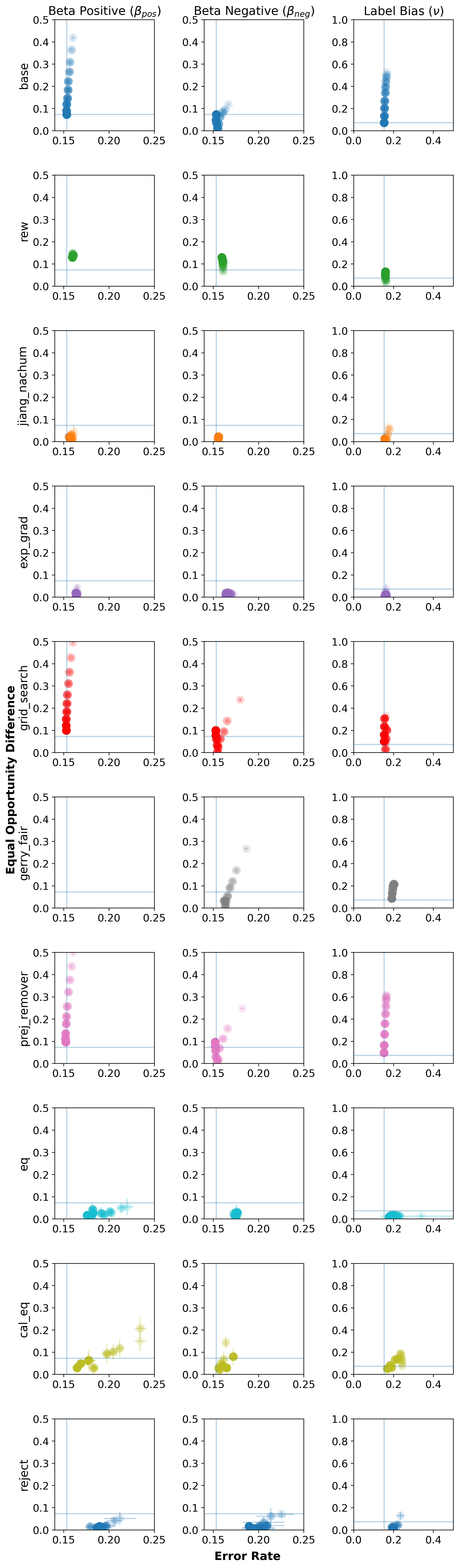}
         \caption{}
     \end{subfigure}
     \begin{subfigure}{0.3\textwidth}
         \includegraphics[width=\textwidth]{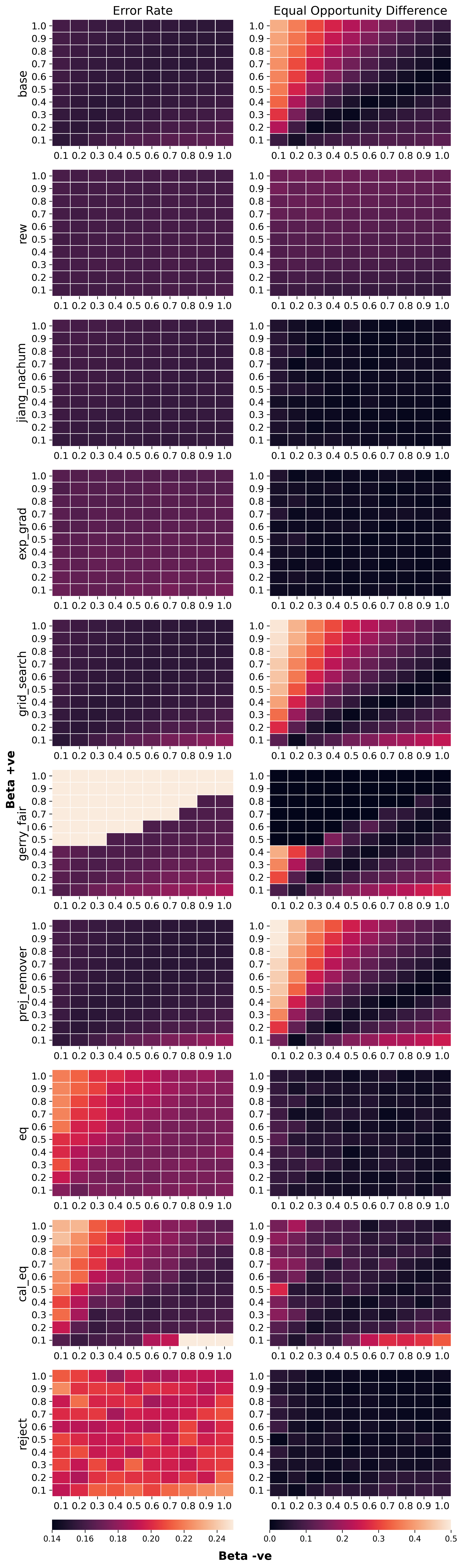}
         \caption{}
     \end{subfigure}
     \begin{subtable}{0.24\textwidth}
            \renewcommand{\arraystretch}{1.35}%
            \centering
            {\notsotiny %
             \begin{tabular}[b]
             {|p{0.7cm}|p{0.3cm}|p{0.75cm}|p{0.75cm}|p{0.75cm}|}
             \hline
              Algo. & & Err. & SPD & EOD \\ \hline
              
              \multirow{2}{*}{base} & lr & \textbf{0.156} $\pm$ 0.003 & 0.171 $\pm$ 0.035 & 0.134 $\pm$ 0.111 \\
              \cline{2-5}
               & svm & \textbf{0.155} $\pm$ \textbf{0.003} & 0.179 $\pm$ 0.036 & 0.172 $\pm$ 0.131 \\
               \hline

               \multirow{2}{*}{rew} & lr & 0.16 $\pm$ \textbf{0.001} & \textbf{0.101} $\pm$ \textbf{0.012} & 0.109 $\pm$ 0.02 \\
              \cline{2-5}
               & svm & \textbf{0.159} $\pm$ \textbf{0.001} & \textbf{0.103} $\pm$ 0.02 & 0.105 $\pm$ 0.029 \\
               \hline

               \multirow{2}{*}{\parbox{0.9cm}{\centering jiang nachum \hspace{0.1cm}}} & lr & \textbf{0.156} $\pm$ \textbf{0.001} & 0.125 $\pm$ \textbf{0.011} & \textbf{0.019} $\pm$ \textbf{0.011} \\
              \cline{2-5}
               & svm & \textbf{0.159} $\pm$ 0.003 & 0.136 $\pm$ \textbf{0.012} & 0.049 $\pm$ \textbf{0.013} \\
               \hline

               \multirow{2}{*}{\parbox{0.9cm}{\centering exp \hspace{0.5cm} grad}} & lr & 0.166 $\pm$ \textbf{0.002} & 0.107 $\pm$ \textbf{0.003} & \textbf{0.016} $\pm$ \textbf{0.008} \\
              \cline{2-5}
               & svm & 0.168 $\pm$ \textbf{0.002} & 0.109 $\pm$ \textbf{0.003} & \textbf{0.017} $\pm$ \textbf{0.008} \\
               \hline

               \multirow{2}{*}{\parbox{0.9cm}{\centering grid search}} & lr & \textbf{0.156} $\pm$ 0.005 & 0.176 $\pm$ 0.048 & 0.169 $\pm$ 0.131 \\
              \cline{2-5}
               & svm & 0.208 $\pm$ 0.041 & \textbf{0.056} $\pm$ 0.067 & 0.051 $\pm$ 0.064 \\
               \hline

               \multirow{2}{*}{\parbox{0.9cm}{\centering gerry fair}} & lr & 0.412 $\pm$ 0.287 & \textbf{0.071} $\pm$ 0.068 & 0.064 $\pm$ 0.087 \\
              \cline{2-5}
               & svm & - & - & - \\
               \hline

               \multirow{2}{*}{\parbox{0.9cm}{\centering prej remover \hspace{0.1cm}}} & lr & 0.156 $\pm$ 0.005 & 0.174 $\pm$ 0.05 & 0.172 $\pm$ 0.135 \\
              \cline{2-5}
               & svm & - & - & - \\
               \hline

               \multirow{2}{*}{eq} & lr & 0.184 $\pm$ 0.013 & \textbf{0.085} $\pm$ 0.019 & 0.038 $\pm$ 0.02 \\
              \cline{2-5}
               & svm & 0.182 $\pm$ 0.013 & \textbf{0.085} $\pm$ \textbf{0.017} & \textbf{0.038} $\pm$ \textbf{0.021} \\
               \hline

               \multirow{2}{*}{cal\_eq} & lr & 0.183 $\pm$ 0.034 & 0.125 $\pm$ 0.075 & 0.088 $\pm$ 0.063 \\
              \cline{2-5}
               & svm & 0.183 $\pm$ 0.036 & 0.123 $\pm$ 0.081 & 0.091 $\pm$ 0.066 \\
               \hline

               \multirow{2}{*}{reject} & lr & 0.199 $\pm$ 0.008 & 0.192 $\pm$ 0.025 & \textbf{0.027} $\pm$ \textbf{0.018} \\
              \cline{2-5}
               & svm & 0.203 $\pm$ 0.008 & 0.191 $\pm$ 0.024 & \textbf{0.027} $\pm$ \textbf{0.02} \\
               \hline
               \multicolumn{5}{c}{}
            \end{tabular}}%
         \caption{}
     \end{subtable}

    \caption{Stability analysis of various fair classifiers on the Adult dataset (Lighter the shade, more the under-representation and label bias): (a) The spread of error rates and Equal Opportunity Difference(EOD) of various fair classifiers as we vary $\beta_{pos}$, $\beta_{neg}$ and label bias separately. The vertical and horizontal reference blue lines denote the performance of an unfair logistic regression model on the original dataset without any under-representation or label bias. (b) Heatmap for Error Rate and EOD across all different settings of $\beta_{pos}$ and $\beta_{neg}$. Darker values denote lower error rates and unfairness. Uniform values across the grid indicate stability to different $\beta_{pos}$ and $\beta_{neg}$. (c) Mean error rates, Statistical Parity Difference (SPD), and Equal Opportunity Difference (EOD) across all $\beta_{pos}$ and $\beta_{neg}$ settings for both kinds of base classifiers (SVM and LR).}
    \label{fig:under_rep_label_bias_adult}
    \end{adjustbox}
\end{figure}

\begin{figure}
     \begin{adjustbox}{minipage=\textwidth}
     \begin{subfigure}{0.3\textwidth}
         \includegraphics[width=\textwidth]{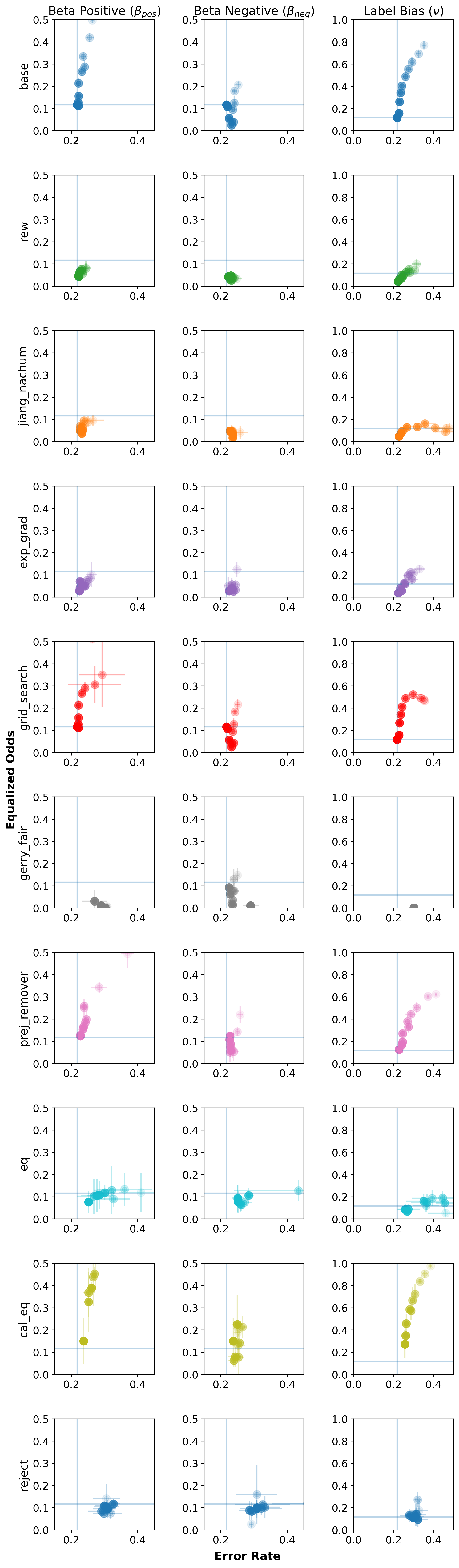}
         \caption{}
     \end{subfigure}
     \begin{subfigure}{0.3\textwidth}
         \includegraphics[width=\textwidth]{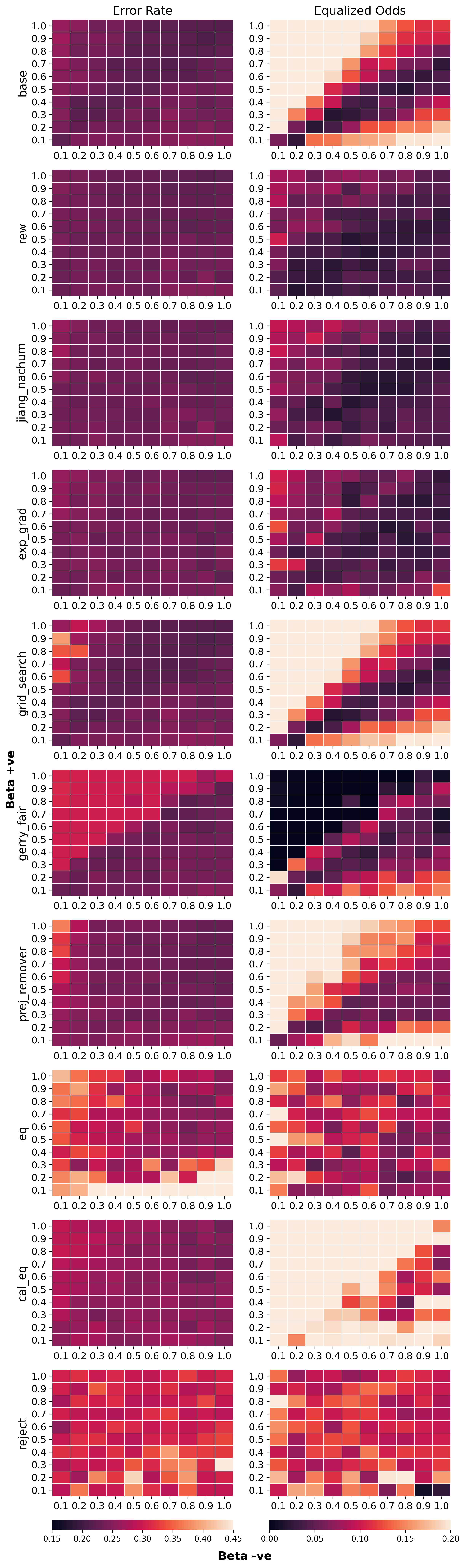}
         \caption{}
     \end{subfigure}
     \begin{subtable}{0.24\textwidth}
            \renewcommand{\arraystretch}{1.35}%
            \centering
            {\notsotiny %
             \begin{tabular}[b]
             {|p{0.7cm}|p{0.3cm}|p{0.75cm}|p{0.75cm}|p{0.75cm}|}
             \hline
              Algo. & & Err. & SPD & EOD \\ \hline
              
              \multirow{2}{*}{base} & lr & \textbf{0.235} $\pm$ 0.012 & 0.156 $\pm$ 0.125 & 0.103 $\pm$  0.11 \\
              \cline{2-5}
               & svm & \textbf{0.235} $\pm$ \textbf{0.011} & 0.17 $\pm$ 0.143 & 0.115 $\pm$ 0.111 \\
               \hline

               \multirow{2}{*}{rew} & lr & \textbf{0.234} $\pm$ \textbf{0.006} & \textbf{0.043} $\pm$ \textbf{0.012} & \textbf{0.018} $\pm$ \textbf{0.009} \\
              \cline{2-5}
               & svm & \textbf{0.234} $\pm$ \textbf{0.007} & 0.071 $\pm$ 0.045 & 0.047 $\pm$ 0.024 \\
               \hline

               \multirow{2}{*}{\parbox{0.9cm}{\centering jiang nachum \hspace{0.1cm}}} & lr & \textbf{0.236} $\pm$ \textbf{0.008} & \textbf{0.038} $\pm$ \textbf{0.012} & \textbf{0.023} $\pm$ \textbf{0.011} \\
              \cline{2-5}
               & svm & \textbf{0.236} $\pm$ \textbf{0.011} & 0.046 $\pm$ \textbf{0.025} & \textbf{0.031} $\pm$ \textbf{0.018} \\
               \hline

               \multirow{2}{*}{\parbox{0.9cm}{\centering exp \hspace{0.5cm} grad}} & lr & 0.239 $\pm$ \textbf{0.008} & 0.045 $\pm$ \textbf{0.022} & 0.027 $\pm$ \textbf{0.014} \\
              \cline{2-5}
               & svm & 0.261 $\pm$ 0.044 & \textbf{0.045} $\pm$ \textbf{0.019} & \textbf{0.034} $\pm$ \textbf{0.015} \\
               \hline

               \multirow{2}{*}{\parbox{0.9cm}{\centering grid search}} & lr & 0.24 $\pm$ 0.027 & 0.153 $\pm$ 0.117 & 0.102 $\pm$ 0.105 \\
              \cline{2-5}
               & svm & 0.291 $\pm$ 0.029 & \textbf{0.026} $\pm$ \textbf{0.028} & \textbf{0.021} $\pm$ \textbf{0.021} \\
               \hline

               \multirow{2}{*}{\parbox{0.9cm}{\centering gerry fair}} & lr & 0.263 $\pm$ 0.031 & \textbf{0.039} $\pm$ 0.038 & \textbf{0.024} $\pm$ 0.024 \\
              \cline{2-5}
               & svm & - & - & - \\
               \hline

               \multirow{2}{*}{\parbox{0.9cm}{\centering prej remover \hspace{0.1cm}}} & lr & 0.247 $\pm$ 0.022 & 0.153 $\pm$ 0.112 & 0.112 $\pm$ 0.114 \\
              \cline{2-5}
               & svm & - & - & - \\
               \hline

               \multirow{2}{*}{eq} & lr & 0.318 $\pm$ 0.074 & 0.057 $\pm$ 0.032 & 0.074 $\pm$ 0.045 \\
              \cline{2-5}
               & svm & 0.336 $\pm$ 0.084 & \textbf{0.045} $\pm$ 0.035 & 0.059 $\pm$ 0.05 \\
               \hline

               \multirow{2}{*}{cal\_eq} & lr & 0.261 $\pm$ 0.013 & 0.255 $\pm$ 0.163 & 0.151 $\pm$ 0.124 \\
              \cline{2-5}
               & svm & 0.255 $\pm$ 0.011 & 0.186 $\pm$ 0.154 & 0.106 $\pm$ 0.111 \\
               \hline

               \multirow{2}{*}{reject} & lr & 0.312 $\pm$ 0.031 & 0.084 $\pm$ 0.038 & 0.098 $\pm$ 0.051 \\
              \cline{2-5}
               & svm & 0.317 $\pm$ 0.037 & 0.089 $\pm$ 0.04 & 0.105 $\pm$ 0.046 \\
               \hline
               \multicolumn{5}{c}{}
            \end{tabular}}%
         \caption{}
     \end{subtable}

    \caption{Stability analysis of various fair classifiers on the Credit dataset (Lighter the shade, more the under-representation and label bias): (a) The spread of error rates and Equalized Odds (EODDS) of various fair classifiers as we vary $\beta_{pos}$, $\beta_{neg}$ and label bias separately. The vertical and horizontal reference blue lines denote the performance of an unfair logistic regression model on the original dataset without any under-representation or label bias. (b) Heatmap for Error Rate and EODDS across all different settings of $\beta_{pos}$ and $\beta_{neg}$. Darker values denote lower error rates and unfairness. Uniform values across the grid indicate stability to different $\beta_{pos}$ and $\beta_{neg}$. (c) Mean error rates, Statistical Parity Difference (SPD), and Equal Opportunity Difference (EOD) across all $\beta_{pos}$ and $\beta_{neg}$ settings for both kinds of base classifiers (SVM and LR).}
    \label{fig:under_rep_label_bias_credit}
    \end{adjustbox}
\end{figure}

\begin{figure}
     \begin{adjustbox}{minipage=\textwidth}
     \begin{subfigure}{0.3\textwidth}
         \includegraphics[width=\textwidth]{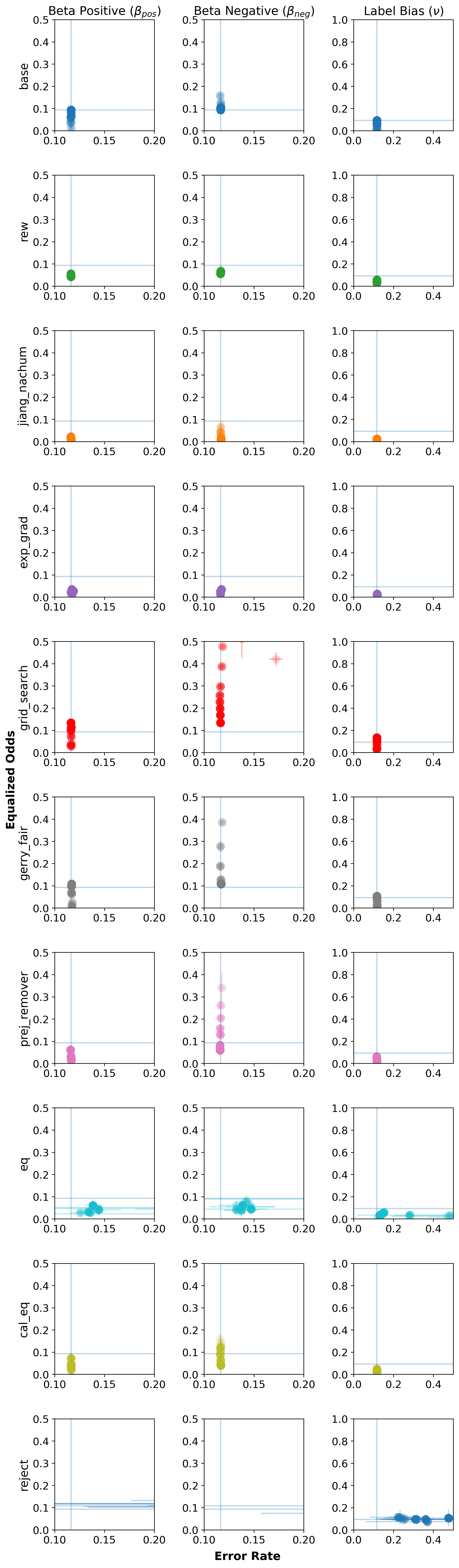}
         \caption{}
     \end{subfigure}
     \begin{subfigure}{0.3\textwidth}
         \includegraphics[width=\textwidth]{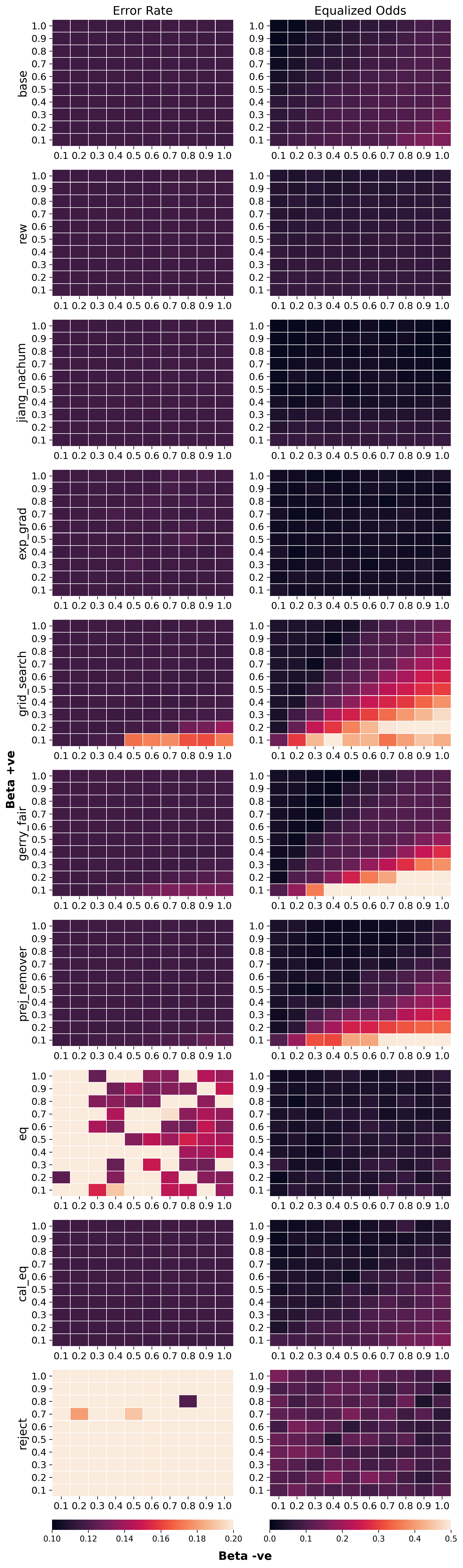}
         \caption{}
     \end{subfigure}
     \begin{subtable}{0.24\textwidth}
            \renewcommand{\arraystretch}{1.3}%
            \centering
            {\notsotiny %
             \begin{tabular}[b]
             {|p{0.7cm}|p{0.3cm}|p{0.75cm}|p{0.75cm}|p{0.75cm}|}
             \hline
              Algo. & & Err. & SPD & EOD \\ \hline
              
              \multirow{2}{*}{base} & lr & \textbf{0.116} $\pm$ \textbf{0.000} & 0.069 $\pm$ 0.021 & 0.12 $\pm$ 0.05 \\
              \cline{2-5}
               & svm & \textbf{0.116} $\pm$ \textbf{0.000} & 0.069 $\pm$ \textbf{0.007} & 0.096 $\pm$ \textbf{0.014} \\
               \hline

               \multirow{2}{*}{rew} & lr & \textbf{0.116} $\pm$ \textbf{0.000} & 0.053 $\pm$ \textbf{0.008} & 0.088 $\pm$ \textbf{0.011} \\
              \cline{2-5}
               & svm & \textbf{0.116} $\pm$ \textbf{0.000} & 0.067 $\pm$ \textbf{0.003} & 0.092 $\pm$ \textbf{0.008} \\
               \hline

               \multirow{2}{*}{\parbox{0.9cm}{\centering jiang nachum \hspace{0.1cm}}} & lr & \textbf{0.116} $\pm$ \textbf{0.000} & \textbf{0.033} $\pm$ 0.016 & \textbf{0.039} $\pm$ 0.031 \\
              \cline{2-5}
               & svm & \textbf{0.116} $\pm$ \textbf{0.000} & 0.05 $\pm$ 0.017 & \textbf{0.051} $\pm$ 0.023 \\
               \hline

               \multirow{2}{*}{\parbox{0.9cm}{\centering exp \hspace{0.5cm} grad}} & lr & 0.117 $\pm$ 0.001 & \textbf{0.023} $\pm$ \textbf{0.007} & \textbf{0.03} $\pm$ \textbf{0.011} \\
              \cline{2-5}
               & svm & 0.124 $\pm$ 0.018 & \textbf{0.022} $\pm$ \textbf{0.006} & \textbf{0.031} $\pm$ \textbf{0.012} \\
               \hline

               \multirow{2}{*}{\parbox{0.9cm}{\centering grid search}} & lr & 0.12 $\pm$ 0.012 & 0.153 $\pm$ 0.14 & 0.255 $\pm$ 0.184 \\
              \cline{2-5}
               & svm & 0.116 $\pm$ 0.001 & \textbf{0.002} $\pm$ 0.01 & \textbf{0.003} $\pm$ 0.015 \\
               \hline

               \multirow{2}{*}{\parbox{0.9cm}{\centering gerry fair}} & lr & 0.117 $\pm$ 0.003 & 0.146 $\pm$ 0.215 & 0.224 $\pm$ 0.238 \\
              \cline{2-5}
               & svm & - & - & - \\
               \hline

               \multirow{2}{*}{\parbox{0.9cm}{\centering prej remover \hspace{0.1cm}}} & lr & 0.116 $\pm$ 0.001 & 0.09 $\pm$ 0.123 & 0.174 $\pm$ 0.175 \\
              \cline{2-5}
               & svm & - & - & - \\
               \hline

               \multirow{2}{*}{eq} & lr & 0.227 $\pm$ 0.125 & \textbf{0.026} $\pm$ \textbf{0.01} & \textbf{0.069} $\pm$ \textbf{0.027} \\
              \cline{2-5}
               & svm & 0.209 $\pm$ 0.093 & \textbf{0.028} $\pm$ 0.01 & 0.067 $\pm$ 0.019 \\
               \hline

               \multirow{2}{*}{cal\_eq} & lr & 0.116 $\pm$ 0.000 & 0.05 $\pm$ 0.028 & 0.099 $\pm$ 0.054 \\
              \cline{2-5}
               & svm & \textbf{0.116} $\pm$ \textbf{0.000} & 0.054 $\pm$ 0.019 & 0.091 $\pm$ 0.025 \\
               \hline

               \multirow{2}{*}{reject} & lr & 0.394 $\pm$ 0.126 & 0.123 $\pm$ 0.025 & 0.118 $\pm$ 0.04 \\
              \cline{2-5}
               & svm & 0.35 $\pm$ 0.115 & 0.116 $\pm$ 0.026 & 0.117 $\pm$ 0.035 \\
               \hline
               \multicolumn{5}{c}{}
            \end{tabular}}%
         \caption{}
     \end{subtable}

    \caption{Stability analysis of various fair classifiers on the Bank dataset (Lighter the shade, more the under-representation and label bias): (a) The spread of error rates and Equalized Odds (EODDS) of various fair classifiers as we vary $\beta_{pos}$, $\beta_{neg}$ and label bias separately. The vertical and horizontal reference blue lines denote the performance of an unfair logistic regression model on the original dataset without any under-representation or label bias. (b) Heatmap for Error Rate and EODDS across all different settings of $\beta_{pos}$ and $\beta_{neg}$. Darker values denote lower error rates and unfairness. Uniform values across the grid indicate stability to different $\beta_{pos}$ and $\beta_{neg}$. (c) Mean error rates, Statistical Parity Difference (SPD), and Equal Opportunity Difference (EOD) across all $\beta_{pos}$ and $\beta_{neg}$ settings for both kinds of base classifiers (SVM and LR).}
    \label{fig:under_rep_label_bias_bank}
    \end{adjustbox}
\end{figure}

Figures \ref{fig:under_rep_label_bias_synthetic}, \ref{fig:under_rep_label_bias_adult}, \ref{fig:under_rep_label_bias_credit} and \ref{fig:under_rep_label_bias_bank} show the scatter plot, heatmap and average results for the Synthetic, Adult, Credit and Bank Marketing datasets respectively. 

\section{Proofs} \label{appndx:proofs}

\subsection{Proof for Theorem 1}
\label{appendix:thm2_proof}

\begin{proof}
We show the proof for binary label $y$ and sensitive attribute $s$. The proof works out in a similar manner for multiple labels and sensitive attributes.

Assume that the subgroup probabilities on $D$ are denoted by $Pr(Y=y, S=s) = p_{ys}$: $p_{00}, p_{01}, p_{10}, p_{11}$.

For the loss $L$, we can write its expectation under distribution $D$ as:
\begin{align*}
& \mathbb{E}_{(X, Y, S) \sim D}[L(f(X), Y)]  = \sum_{y, s} Pr(Y=y, S=s)~ \mathbb{E}_{X | Y=y, S=s}[L(f(X),y)]
\end{align*}
Let $m = \min\{p_{00}, p_{01}, p_{10}, p_{11}\}$, $M = \max\{p_{00}, p_{01}, p_{10}, p_{11}\}$, and $\alpha=m/M$.  

For the rest of the proof, we will assume an under-representation bias on $(Y = 1, S=0)$ subgroup with bias $\beta$. The proof for bias on $(Y = 0, S=0)$ subgroup will work out similarly.

We denote by $D_{\beta}$ the distribution induced because of the under-representation bias on $(Y = 1, S=0)$ subgroup, where $p'_{10} \propto \beta p_{10}$, and $p'_{ys} \propto p_{ys}$ for rest of the subgroups.

For the reweighing loss $L'$, the reweighing weight $W_{ys}$ is given by the following expression ~\cite{kamiran2012data}:
\[
W_{ys} = \frac{Pr(Y'=y)~ Pr(S'=s)}{Pr(Y'=y, S'=s)}.
\]
Now we can write
\begin{align*}
& \mathbb{E}_{(X', Y', S') \sim D_{\beta}}[L'(f(X'), Y')] = \sum_{y, z} Pr(Y'=y, S'=s)~ \mathbb{E}_{X'| Y'=y, S'=z}[L'(f(X'),y)] \\
& = \sum_{y, s} Pr(Y'=y, S'=s)~ \mathbb{E}_{X'| Y'=y, S'=s}[W_{ys} L(f(X'), y)] \\
& = \sum_{y, s} Pr(Y'=y, S'=s)~ W_{ys}~ \mathbb{E}_{X'| Y'=y, S'=s}[L(f(X'), y)] \\
& = \sum_{y, s} Pr(Y'=y)~ Pr(S'=s)~ \mathbb{E}_{X'| Y'=y, S'=s}[L(f(X'), y)].
\end{align*}
Thus, we have 
\begin{align*}
& \mathbb{E}_{D_{\beta}}[L'] = \mathbb{E}_{(X', Y', S') \sim D_{\beta}}[L'(f(X'), Y')] \\
& = \sum_{y, z} Pr(Y'=y)~ Pr(S'=s)~ \mathbb{E}_{X'| Y'=y, S'=s}[L(f(X'), y)] \\
& = \sum_{y, z} Pr(Y'=y)~ Pr(S'=s)~ \mathbb{E}_{X| Y=y, S=s}[L(f(X), y)],
\end{align*}
because $X'| Y'=y, S'=s$ in $D_{\beta}$ remains the same as $X| Y=y, S=s$ in $D$ even after injecting under-representation.

For the distribution $D_{\beta}$, assume that the normalization constant for the subgroup probabilities  is $N' = \sum_{Y'=y, S'=s} p_{ys} = p_{00} + p_{01} + \beta p_{10} + p_{11}$.

We can write down individual label and sensitive attribute probabilities as follows in $D_{\beta}$: $\Pr(Y'=0) = (p_{00} + p_{01})/N'$, $Pr(Y'=1)=(\beta p_{10} + p_{11})/N'$, $Pr(S'=0)=(p_{00} + \beta p_{10})/N'$ and $Pr(S'=1)=(p_{01} + p_{11})/N'$.

We will now attempt to lower and upper bound 
\[
\frac{Pr(Y'=y)~ Pr(S'=s)}{Pr(Y=y, S=s)}
\]
in terms of $\alpha$, for any $Y,S \in \{0, 1\}$ and any $\beta \in (0,1]$. Let $N = 1 = p_{00} + p_{01} + p_{10} + p_{11}$. For example, for $Y=0, S=0$ we get
\[
\frac{\alpha}{2} \leq \frac{p_{00}+p_{01}}{N} \leq \frac{(p_{00}+p_{01})}{N'}~ \frac{(p_{00}+\beta p_{10})}{N'}~ \frac{N}{p_{00}} \leq \frac{N}{p_{00}} \leq \frac{4}{\alpha},
\]
using $N' \leq N$, $p_{00} \leq p_{00} + \beta p_{10}$ and the definition of $m, M$ and $\alpha$. Similarly working out the upper and lower bounds for each of the four probability terms in the expected reweighed loss expression above, we get a common bound 
\[
\frac{\alpha^{2}}{4} \leq \frac{Pr(Y'=y)~ Pr(S'=s)}{Pr(Y=y, S=s)} \leq \frac{4}{\alpha},
\]
for any $y,s \in \{0, 1\}$. Now we can plug in these bounds in the following expression for $\mathbb{E}_{D}[L]$.
\begin{align*}
& \mathbb{E}_{D}[L] = \mathbb{E}_{(X, Y, S) \sim D}[L(f(X), Y)] \\
& = \sum_{y, s} Pr(Y=y, S=s)~ \mathbb{E}_{X| Y=y, S=s}[L(f(X), y)]    
\end{align*}

Using the sandwich bound with the final form of $\mathbb{E}_{D_{\beta}}[L']$, we get: 
\[
\frac{\alpha^{2}}{4}~ \mathbb{E}_{D}[L] \leq \mathbb{E}_{D_{\beta}}[L'] \leq \frac{4}{\alpha}~ \mathbb{E}_{D}[L].
\]
\end{proof}

\subsection{Proof of Theorem 2}
\label{appendix:thm3_proof}

Before proving Theorem 1, we prove a small proposition that will be useful for later results.

\begin{proposition} \label{prop:hoeff}
Let L be a non-negative bounded loss between $[0,B]$. Let $\hat{f}$ be the empirical risk minimizing classifier with loss $L$ with $N$ samples: $\hat{f} = \argmin_{f \in \mathcal{F}} \frac{1}{N} \sum_{i=1}^{N} L(f(x_{i}, y_{i}))$, where $\{(x_{1}, y_{1}), (x_{2}, y_{2}), ..., (x_{N}, y_{N})\}$ are i.i.d. samples from $D$. Let $f^{*} = \argmin_{f \in \mathcal{F}} \mathbb{E}_{(x,y) \sim D} [L(f(x), y)]$. Then with probability of at least $1 - \delta$, we have
\[
\mathbb{E}_{D}[L(\hat{f})] \leq B~ \sqrt{\frac{\log(2/\delta)}{2N}} + \mathbb{E}_{D}[L(f^{*})].
\]
\end{proposition}

\begin{proof}
The proof follows directly from the proof of Thm. 1 in \cite{zhu2021rich}, and a general form of Hoeffding's inequality for any non-negative bounded loss\footnote{http://cs229.stanford.edu/extra-notes/hoeffding.pdf}.
\end{proof}

We are now ready to prove Theorem 2.

\begin{proof}
Using the results from Theorem 1, for any $g$ and $\beta$, we can write
\[
\frac{\alpha^{2}}{4}~ \mathbb{E}_{D}[L(g)] \leq \mathbb{E}_{D_\beta}[L'(g)] \leq \frac{4}{\alpha}~ \mathbb{E}_{D}[L(g)].
\]
For $\hat{g}_{\beta}$, we can use the lower bound from Theorem 1 to get
\begin{align} \label{eq:L-to-L'}
\mathbb{E}_{D}[L(\hat{g}_{\beta})] \leq \frac{4}{\alpha^{2}}~ \mathbb{E}_{D_\beta}[L'(\hat{g}_{\beta})].
\end{align}
We now assume that $L$ is a bounded loss between $[0,1]$. if not, we can rescale it to lie in $[0, 1]$. Using $m = \min\{p_{00}, p_{01}, p_{10}, p_{11}\}$, $M = \max\{p_{00}, p_{01}, p_{10}, p_{11}\}$, and $\alpha=m/M$, we can show that the reweighing factor
\[
W_{ys} = \frac{Pr(Y'=y)~ Pr(S'=s)}{Pr(Y'=y, S'=s)} \leq \frac{4}{\alpha \beta}.
\]
for all $y, s \in \{0, 1\}$. This can be done by using the upper bound from Theorem 1 and the fact that in the worst case ($Y=1, S=0$ here),  $\frac{Pr(Y=y,S=s)}{Pr(Y'=y, S'=s)} = \frac{1}{\beta}$.

Thus, the reweighed loss $L'$ is bounded between $[0, 4/(\alpha \beta)]$. Now, using Proposition \ref{prop:hoeff} and that $\hat{g}$ is the empirical minimizer of the reweighed loss $L'$ on $N$ samples from the biased distribution $D_{\beta}$, we obtain that 
\[
\mathbb{E}_{D_{\beta}}[L'(\hat{g}_{\beta})] \leq \frac{4}{\alpha \beta}~ \sqrt{\frac{\log(2/\delta)}{2N}} + \mathbb{E}_{D_{\beta}}[L'(g^{*})].
\]
where $g^{*}$ is the Bayes optimal classifier for the reweighed loss $L'$ on the biased distribution $D_{\beta}$. Now let $f^{*}$ be the Bayes optimal classifier for the loss $L$ on the original distribution $D$. Thus, $\mathbb{E}_{D_{\beta}}[L'(g^{*})] \leq \mathbb{E}_{D_{\beta}}[L'(f^{*})]$, and we get
\begin{align} \label{eq:L'-gen}
\mathbb{E}_{D_{\beta}}[L'(\hat{g}_{\beta})] \leq \frac{4}{\alpha \beta}~ \sqrt{\frac{\log(2/\delta)}{2N}} + \mathbb{E}_{D_{\beta}}[L'(f^{*})].
\end{align}
Combining \eqref{eq:L-to-L'} and \eqref{eq:L'-gen}, we have
\begin{align*}
\mathbb{E}_{D}[L(\hat{g}_{\beta})] 
& \leq \frac{4}{\alpha^{2}} \left(\frac{4}{\alpha \beta}~ \sqrt{\frac{\log(2/\delta)}{2N}} + \mathbb{E}_{D_{\beta}}[L'(f^{*})]\right) \\
& = \frac{16}{\alpha^{3} \beta}~ \sqrt{\frac{\log(2/\delta)}{2N}} + \frac{4}{\alpha^{2}}~ \mathbb{E}_{D_{\beta}}[L'(f^{*})] \\
& \leq \frac{16}{\alpha^{3} \beta}~ \sqrt{\frac{\log(2/\delta)}{2N}} + \frac{16}{\alpha^{3}}~ \mathbb{E}_{D}[L(f^{*})],
\end{align*}
where we use the upper bound in Theorem 1 to get the last inequality. That completes the proof of Theorem 2.
\end{proof}

\end{document}